\documentclass[lettersize,journal]{IEEEtran}
\usepackage{amsmath,amsfonts}
\usepackage{algorithm}
\usepackage{array}
\usepackage[caption=false,font=normalsize,labelfont=sf,textfont=sf]{subfig}
\usepackage{textcomp}
\usepackage{stfloats}
\usepackage{booktabs}
\usepackage{xspace}
\usepackage{enumitem}
\usepackage{url}
\usepackage{multirow}
\usepackage{verbatim}
\usepackage{graphicx}
\usepackage{cite}
\usepackage{algpseudocode}
\usepackage{amssymb}
\usepackage{xcolor}
\usepackage{booktabs}
\usepackage{hyperref}
\definecolor{utpColor}{RGB}{0, 90, 170}
\definecolor{staColor}{RGB}{180, 60, 0}

\newcommand{\method}{\text{FactoST-v2}\xspace}
\newcommand{\boldres}[1]{{\textbf{\textcolor{red}{#1}}}}
\newcommand{\secondres}[1]{{\underline{\textcolor{blue}{#1}}}}

\begin{document}

\title{Learning to Factorize and Adapt: A Versatile Approach Toward Universal Spatio-Temporal Foundation Models}

\author{Siru Zhong, Junjie Qiu, Yangyu Wu, Yiqiu Liu, Yuanpeng He, \\ Zhongwen Rao, Bin Yang, Chenjuan Guo, Hao Xu, and Yuxuan Liang
\thanks{This article is an extension of our preliminary version presented at the 39th Conference on Neural Information Processing Systems (NeurIPS 2025)~\cite{zhonglearning}.}
\thanks{Siru Zhong, Junjie Qiu, Yangyu Wu, Yiqiu Liu, and Yuxuan Liang are with The Hong Kong University of Science and Technology (Guangzhou). Corresponding author: Yuxuan Liang (yuxliang@outlook.com)}%
\thanks{Yuanpeng He is with Peking University.}%
\thanks{Zhongwen Rao and Hao Xu are with Huawei 2012 Laboratories.}%
\thanks{Bin Yang and Chenjuan Guo are with East China Normal University.}%
}



\maketitle

\begin{abstract}
Spatio-Temporal (ST) Foundation Models (STFMs) promise cross-dataset generalization, yet joint ST pretraining is computationally expensive and grapples with the heterogeneity of domain-specific spatial patterns. Substantially extending our preliminary conference version~\cite{zhonglearning}, we present \method, an enhanced factorized framework redesigned for full weight transfer and arbitrary-length generalization. \method decouples universal temporal learning from domain-specific spatial adaptation. The first stage pretrains a minimalist encoder-only backbone using randomized sequence masking to capture invariant temporal dynamics, enabling probabilistic quantile prediction across variable horizons. The second stage employs a streamlined adapter to rapidly inject spatial awareness via meta adaptive learning and prompting. Comprehensive evaluations across diverse domains demonstrate that \method achieves state-of-the-art accuracy with linear efficiency—significantly outperforming existing foundation models in zero-shot and few-shot scenarios while rivaling domain-specific expert baselines. This factorized paradigm offers a practical, scalable path toward truly universal STFMs. Code is available at \url{https://github.com/CityMind-Lab/FactoST.git}.
\end{abstract}


\section{Introduction}


\IEEEPARstart{S}{patio-temporal} (ST) patterns govern the evolution of dynamic signals across complex physical systems—ranging from the daily periodicity of traffic flows on road networks to the diffusive dispersion of air pollutants across citywide sensor networks. Unraveling these intricate patterns is fundamental to forecasting and decision support, enabling proactive anticipation in critical applications in traffic, energy, environment and ubiquitous domains \cite{zheng2014urban,faghmous2014spatio,wang2020deep,chen2024semantic,zhang2024predicting,feng2024spatio,zhou2024navigating,wang2025airradar,han2025adamove,zhong2025multimodal,ruan2025cross,liu2025towards,zou2025fine,tianair,zheng2025multi,zhu2025multiscale,wang2025gravity}.

\begin{figure}[!t]
    \centering
    \includegraphics[width=\linewidth]{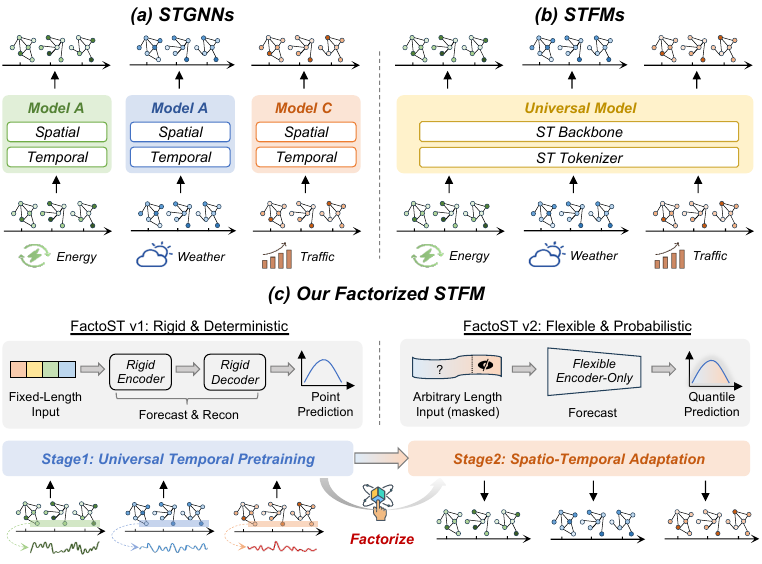}
    \caption{\textbf{Evolution of Spatio-Temporal Modeling Paradigms.} (a) \textbf{Traditional STGNNs}: Coupled modeling tailored for specific graphs. (b) \textbf{Existing STFMs}: Joint pretraining on heterogeneous graphs, limited by topological conflicts. (c) \textbf{Our Factorized STFM (FactoST)}: The upper panel highlights the \textbf{architectural evolution} from the rigid, deterministic design of FactoST-v1 (Left) to the flexible, probabilistic backbone of FactoST-v2 (Right). The lower panel illustrates the core \textit{Factorized Paradigm}—decoupling universal temporal learning (Stage 1) from spatial adaptation (Stage 2).}
    \vspace{-1em}
    \label{motivation}
\end{figure}

In deep learning practice, the modeling of these patterns has traditionally relied on Spatio-Temporal Graph Neural Networks (STGNNs)~\cite{wang2020deep,liu2023we,sahili2023spatio,jin2023spatio,jin2024survey, Wang2025stgcn}. As depicted in Figure \ref{motivation}(a), these task-specific models typically adopt a \emph{coupled} modeling paradigm: they interweave a temporal module (e.g., RNN~\cite{bai2020adaptive,qiu2018spatio}, CNN~\cite{yu2017spatio,wu2019graph,kong2020stgat}, or Attention~\cite{liang2018geoman,zhang2019spatial,guo2019attention}) to extract sequential dynamics with a spatial module (e.g., GCN~\cite{kipf2016semi,li2017diffusion}) to propagate information along a predefined graph. While effective within specific domains, this coupled design inherently binds the model to fixed spatial topologies, limiting its ability to generalize patterns across datasets.

Inspired by the success of Foundation Models (FMs) in language and vision~\cite{openai2023gpt4,awais2025foundation,chowdhery2023palm}, recent research has pivoted towards \textbf{Spatio-Temporal Foundation Models (STFMs)}~\cite{liang2025foundation,goodge2025spatio,huang2025foundation,fang2025unraveling,zhang2024urban}. The prevailing paradigm (Figure \ref{motivation}(b)) aims to learn a monolithic model through \emph{joint ST pretraining} on massive corpora, following a core principle: \emph{Pretrain a single model on diverse ST corpora (e.g., climate, traffic, energy) and adapt it to unseen datasets in a zero-shot or few-shot fashion}. Such cross-domain pretraining equips STFMs with broad ST knowledge and cross-dataset generalization, often outperforming task-specific STGNNs when labeled data is scarce~\cite{liu2024unitime,li2024opencity,yuan2025urbandit}. However, we argue that this ``joint learning'' approach faces a fundamental \textbf{Pattern Mismatch} challenge arising from the distinct nature of spatial and temporal laws:

\begin{table}[t!]
    \centering
    \caption{Comparison between the conference version (FactoST) and this journal version (\method). \method simplifies the architecture to achieve full weight transfer and arbitrary-length generalization.}
    \label{tab:diff_table}
    \resizebox{\columnwidth}{!}{%
    \begin{tabular}{l|l|l}
    \toprule
    \textbf{Aspect} & \textbf{Conf. Version (FactoST)} & \textbf{Journal Version (\method)} \\
    \midrule
    \textbf{Backbone Design} & Enc-Decoder (Fixed Horizon) & \textbf{Enc-Only (Arbitrary Horizon)} \\
    \midrule
    \textbf{Pretraining Obj.} & Hybrid (Reconstruction + Pred.) & \textbf{Pure (Quantile Prediction)} \\
    \midrule
    \textbf{Adaptation Module} & Hierarchical Alignment (HDA) & \textbf{Streamlined Alignment (DSPA)} \\
    \midrule
    \textbf{Uncertainty} & Deterministic Point Estimate & \textbf{Probabilistic Quantile} \\
    \midrule
    \textbf{Param. Transfer} & Partial Transfer (Decoder Gap) & \textbf{Full Weight Transfer} \\
    \midrule
    \textbf{Evaluation Scope} & Basic Few-shot Performance & \textbf{Scaling Laws, Zero/Full-shot} \\
    \bottomrule
    \end{tabular}%
    }
    \vspace{-1em}
\end{table}
\begin{itemize}[leftmargin=*]
    \item \textbf{Domain-Invariant Temporal Patterns:} Temporal dynamics \cite{ren2023deep} (e.g., trends, seasonality) often share universal 1-D physical structures across domains. A seasonal cycle in electricity load shares spectral similarities with traffic flow.
    
    \item \textbf{Domain-Specific Spatial Patterns:} Spatial correlations are tied to domain-specific topologies (e.g., power grids \cite{liao2021review} vs. road networks \cite{li2017diffusion}) and physical distances, which are difficult to unify into a single manifold. For example, neighbouring air-quality stations in Beijing exhibit short-range diffusion dynamics \cite{yi2018deep}, whereas teleconnection effects dominate climate indices across the Pacific Ocean \cite{ham2019deep}.
\end{itemize}

Moreover, existing STFMs~\cite{yuan2024unist, li2024opencity} that force the joint modeling of these diverging spatio-temporal patterns suffer from quadratic complexity. The memory and time footprints grow quadratically with sequence length or graph size, often leading to the over-squashing problem at scale~\cite{marisca2025over} (\autoref{tab:model_comparison}).

In this paper, we address these challenges by proposing a \textbf{Pattern Factorization Hypothesis}: \emph{Effective ST generalization requires decoupling domain-invariant temporal dynamics from domain-specific spatial contexts.} Based on this hypothesis, we introduce \textbf{\method}, a \textit{versatile} framework that factorizes STFM learning into two lightweight stages. As illustrated in the lower panel of Figure \ref{motivation}(c), instead of learning a heavy, coupled representation, \method exploits the asymmetry of ST patterns: it first distills universal temporal laws into a shared backbone, and then injects domain-specific spatial patterns via a lightweight adapter. Conceptually, \method redefines the ``STGNN'' in the era of foundation models, establishing ST factorization at scale.

\vspace{0.3em}
\noindent\textbf{Universal Temporal Pretraining (UTP):} The first stage learns general temporal knowledge across domains. Unlike prior designs that are often complex (e.g., Encoder-Decoder with auxiliary reconstruction tasks) or inflexible (e.g., fixed input-output horizons), we propose a minimalist \textit{Encoder-Only} backbone. By incorporating a learnable \texttt{[REG]} token to isolate context from prediction horizons, combined with a \textit{randomized sequence masking} strategy and frequency-preserving Rotary Positional Embeddings (RoPE), the backbone effectively learns an \emph{arbitrary input-output length mapping}. This design shift unlocks \textit{versatility} in sequence modeling: it seamlessly accommodates \textit{variable-length generalization} and \textit{probabilistic quantile modeling} to quantify inherent uncertainties. Overall, this stage is graph-agnostic, lightweight, scalable, and fully transferable, allowing the model to capture universal sequential dependencies and uncertainty distributions without being contaminated by domain-specific spatial noise.

\vspace{0.3em}
\noindent\textbf{Spatio-Temporal Adaptation (STA):} For a target dataset, the second stage adapts the UTP backbone to inject ST awareness. Benefiting from the streamlined encoder-only backbone, the adaptation process is significantly simplified. We discard the heavy hierarchical alignment module used in our previous version. Instead, we directly inject domain specificity via \textit{learnable prompt tokens} at the input stage. The adapter enriches temporal features with ST identifiers and dynamically reweights them via context relevance, followed by a small memory replay buffer to stabilize training. This streamlined mechanism allows for direct and efficient interaction with the pretrained backbone, ensuring \emph{full weight transfer} of the pretrained temporal knowledge while minimizing the parameters required to model spatial correlations.

Building upon our conference version FactoST~\cite{zhonglearning}—which pioneered the factorized paradigm for STFMs—this journal version (\method) executes a comprehensive and strategic architectural shift to maximize the model's \textit{versatility} across diverse regimes and tasks. We identify two critical limitations in the preliminary work: \textbf{First}, the structural rigidity of the Encoder-Decoder architecture imposed fixed input-output lengths, limiting generalization to varying downstream horizons. Moreover, the decoder components were often discarded during adaptation, leading to \textit{partial weight transfer}. \textbf{Second}, the reliance on auxiliary reconstruction tasks introduced \textit{optimization redundancy}, increasing complexity without guaranteeing better forecasting alignment. In \method, we resolve these issues by transitioning to a \textit{Probabilistic Encoder-Only} framework. As visualized in the upper panel of Figure \ref{motivation}(c), this architectural evolution eliminates structural rigidity, bringing substantial upgrades in two dimensions:

\begin{enumerate}[leftmargin=*]
    \item \textbf{Methodological Upgrades:} We replace the rigid encoder-decoder with a flexible encoder-only backbone. This shift eliminates structural gaps between the two stages, ensuring \textit{100\% parameter transfer} and enabling \textit{arbitrary-length generalization}. Furthermore, we upgrade the objective from deterministic estimation to \textit{probabilistic quantile prediction}, enabling robust uncertainty quantification.

    \item \textbf{Experimental Expansions:} We significantly widen the evaluation scope to address five research questions. Beyond the original few-shot setting, we validate \method across \textit{zero-shot} and \textit{full-shot} scenarios (RQ1). We conduct granular ablation studies to verify component efficacy (RQ2), systematically investigate \textit{scaling laws} regarding data and model size (RQ3), analyze \textit{interpretability} and \textit{universality} (RQ4), and perform rigorous \textit{efficiency profiling} (RQ5).
\end{enumerate}

\begin{table}[b!]
    \centering
    \caption{Comparison of ST modeling paradigms. \textbf{Key Advantage:} Unlike Coupled approaches that suffer from quadratic complexity and topology constraints, our \textbf{Factorized} paradigm achieves linear complexity and universal transferability. (Complexity Notation: $N$: nodes, $T$: horizon, $P$: patch size).}
    \label{tab:model_comparison}
    
    \renewcommand{\arraystretch}{1.1} 
    \setlength{\tabcolsep}{3.5pt}
    
    \footnotesize 
    \resizebox{\linewidth}{!}{%
    \begin{tabular}{l|c|c|c}
    \toprule
    \textbf{Dimension} & \textbf{Traditional STGNNs} & \textbf{Joint STFMs} & \textbf{Factorized STFM} \\
    \midrule
    \textbf{Representative Models} & 
    \textit{DCRNN~\cite{li2017diffusion}, GWNet~\cite{wu2019graph}} & 
    \textit{UniST~\cite{yuan2024unist}, OpenCity~\cite{li2024opencity}} & 
    \textbf{\textit{FactoST~\cite{zhonglearning}, FactoST-v2}} \\
    \midrule
    
    \textbf{Learning Paradigm} & 
    Coupled (End-to-End) & 
    Coupled (Joint Pretrain) & 
    \textbf{Decoupled (UTP + STA)} \\
    
    \textbf{Pattern Encoding} & 
    Topology-Bound & 
    Topology-Aware & 
    \textbf{Topology-Agnostic} \\
    
    \textbf{Spatial Dependency} & 
    Static Graph & 
    Dense / Grid Attention & 
    \textbf{Adaptive Prompting} \\
    
    \textbf{Transferability} & 
    None ($\times$) & 
    Limited (Seen Domains) & 
    \textbf{Universal (Zero-Shot)} \\
    
    \midrule
    
    \textbf{Time Complexity} & 
    $\mathcal{O}(T(N+E))$ & 
    $\mathcal{O}(N^2 T^2)$ \textcolor{red}{(High)} & 
    \textbf{$\mathcal{O}(N (T/P)^2)$ \textcolor{blue}{(Low)}} \\
    
    \textbf{Memory Footprint} & 
    Moderate & 
    High (Quadratic) & 
    \textbf{Low (Linear)} \\
    
    \textbf{Scalability} & 
    Limited & 
    Bottlenecked & 
    \textbf{Highly Scalable} \\
    
    \bottomrule
    \end{tabular}%
    }
\end{table}

In summary, the main contributions of this paper are:

\begin{itemize}[leftmargin=*]
    \item \textbf{Innovative Paradigm:} We propose a truly scalable factorized framework for STFMs that decouples universal temporal pretraining from lightweight ST adaptation. This design circumvents the quadratic complexity of joint pretraining, paving a practical path toward truly \textit{versatile} STFMs featuring \textit{architecturally agnostic, plug-and-play} components.

    \item \textbf{Minimalist Design:} We present a streamlined encoder-only backbone supporting arbitrary-length probabilistic modeling. This minimalist design facilitates full weight transfer and robust generalization, proving that complexity is not a prerequisite for high performance.

    \item \textbf{Comprehensive Evaluation:} We establish a rigorous benchmark covering various settings and analysis. Extensive experiments demonstrate that \method achieves state-of-the-art accuracy with exceptional efficiency, verifying its potential as a general-purpose ST backbone that unifies \textit{zero-shot, few-shot, and full-shot} regimes.
\end{itemize}

Notably, we adopt forecasting as the primary evaluation protocol, as it rigorously tests the model's grasp of ST dynamics and aligns with established benchmarks for fair comparison.

\section{Preliminary}

\subsection{Formulation}

\begin{figure}[h!]
    \centering
    \includegraphics[width=\linewidth]{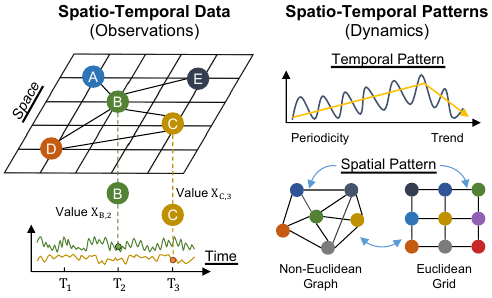}
    \caption{\textbf{Illustration of Spatio-Temporal Modeling.} The figure characterizes ST modeling from two perspectives: (Left) \textbf{Spatio-Temporal Data (Observations):} Visualizing the evolution of dynamic signals ($\mathbf{X}$) across spatial nodes and time steps. (Right) \textbf{Spatio-Temporal Patterns (Dynamics):} Depicting the underlying mechanisms, where \textit{Temporal Patterns} exhibit universal properties over time (e.g., periodicity, trends), while \textit{Spatial Patterns} manifest in heterogeneous topologies—ranging from \textbf{Non-Euclidean Graphs} (e.g., irregular traffic networks) to \textbf{Euclidean Grids} (e.g., regular raster data).}
    \label{fig:st_pattern}
\end{figure}

\vspace{0.3em}
\noindent\textbf{Spatio-Temporal (ST) Data Modeling.}
We formulate the ST modeling problem as learning the evolution of dynamic signals over complex spatial structures. Let $\mathcal{D} = \{(\mathbf{X}, \mathbf{A}, \mathbf{M})\}$ denote a ST dataset. ((1) \textbf{Observations $\mathbf{X}$}: Let $\mathcal{V}=\{v_1, \dots, v_N\}$ be a set of $N$ spatial nodes (e.g., sensors, regions). As shown in Figure \ref{fig:st_pattern} (Left), at each time step $t$, the system states are observed as $\mathbf{X}_t \in \mathbb{R}^{N \times D}$, where $D$ is the feature dimension. For a history window $L$, the input tensor is $\mathcal{X}_{t-L+1:t} \in \mathbb{R}^{N \times L \times D}$. (2) \textbf{Spatial Topology $\mathbf{A}$}: The structural interactions are encapsulated by an adjacency matrix $\mathbf{A} \in \mathbb{R}^{N \times N}$, which may be predefined (e.g., road distance) or learned latent correlations. \textit{Crucially, distinct domains possess highly heterogeneous graph structures.} As illustrated in the spatial patterns of Figure \ref{fig:st_pattern}, these topologies range from irregular, non-Euclidean graphs (e.g., traffic networks) to regular, Euclidean grids (e.g., air quality matrices). This structural variability creates a fundamental barrier for unified modeling. (3) \textbf{Node Metadata $\mathbf{M}$}: Domain-specific semantics are provided by metadata $\mathbf{M} \in \mathbb{R}^{N \times D_m}$ (e.g., geo-coordinates), serving as static identifiers for spatial contexts. This formulation unifies diverse data modalities, including multivariate time series (where $\mathbf{A}$ is implicit)~\cite{zhou2021informer,zhou2022fedformer,du2020multivariate,wang2024deep} and grid-based raster data~\cite{zhang2017deep,yuan2024unist,atluri2018spatio,guo2019deep}, while highlighting the separation between spatial specificity and temporal universality (Figure \ref{fig:st_pattern}).

\vspace{0.3em}
\noindent\textbf{Objective of ST Foundation Models (STFMs).}  
Unlike task-specific models trained on a single dataset $\mathcal{D}_{target}$, STFMs aim to learn a universal representation function $\Phi(\cdot)$ from a massive collection of source domains $\mathbb{S} = \{\mathcal{D}_1, \dots, \mathcal{D}_{K}\}$ that generalizes to unseen target domains $\mathbb{T}$. Standard STFMs typically adopt \emph{forecasting} as the primary proxy task~\cite{liang2025foundation,yuan2024unist,goodge2025spatio}: given historical observations $\mathcal{X} \in \mathbb{R}^{N \times L \times D}$, the model predicts the future horizon $\mathbf{Y} \in \mathbb{R}^{N \times H \times D_{out}}$. Ideally, an STFM should satisfy two critical properties:\begin{itemize}[leftmargin=*]
    \item \textbf{\textit{Topology-Agnostic Versatility}}: The model must handle variable input lengths $L$, prediction horizons $H$, and, most importantly, diverse spatial topologies (varying $N$ and $\mathbf{A}$) across domains without structural reconfiguration. It should support Zero-shot and Few-shot generalization on $\mathbb{T}$.
    \item \textbf{\textit{Parameter-Efficient Adaptation}}: Adapting the universal representations to a specific target domain $\mathcal{D}_{tgt} \in \mathbb{T}$ should require minimal computational overhead (e.g., freezing the backbone and tuning lightweight adapters), surpassing the efficiency of training domain-experts from scratch.
\end{itemize}

\subsection{Related Work}
\noindent\textbf{ST Graph Neural Networks (STGNNs).}
STGNNs \cite{ying2021transformers,sahili2023spatio} constitute the cornerstone of modern ST pattern mining, widely adopted for forecasting, anomaly detection, and imputation tasks~\cite{sun2024survey,jin2023spatio,jin2024survey,gao2022generative,alam2022survey}. The dominant paradigm relies on \emph{structural coupling}: intertwining temporal encoders (e.g., RNNs~\cite{li2017diffusion,huang2023mapredrnn}, TCNs~\cite{wu2019graph,wu2020connecting}) to capture sequential dynamics with spatial modules (e.g., GCNs~\cite{kipf2016semi,zhong2022spatio}, GATs~\cite{zhang2019spatial,song2022st}) to propagate information across graph topologies. While early works relied on predefined static graphs, recent advances have focused on \textit{Graph Structure Learning (GSL)} to infer latent dynamic correlations from data. For instance, Feng \emph{et al.}~\cite{feng2022adaptive} and Shang \emph{et al.}~\cite{shang2021discrete} proposed adaptive mechanisms to learn evolving adjacencies, effectively capturing time-varying spatial dependencies. This coupled design—whether implemented in stacked layers~\cite{yu2017spatio,zhao2019t} or tightly integrated pipelines~\cite{li2017diffusion}—has proven effective across domains like traffic forecasting (e.g., \textsc{BigST}~\cite{han2024bigst}, a linear complexity STGNN for large-scale networks), energy management~\cite{agoua2017short,tascikaraoglu2018evaluation}, and environmental monitoring~\cite{liang2023airformer,davidson2022st}. Despite these advances, most models are trained from scratch per dataset, \emph{inherently limiting cross-domain reuse and falling short of the ``train once, adapt everywhere'' paradigm}.

\vspace{0.3em}
\noindent\textbf{ST Representation Learning.} 
To reduce dependency on massive labeled data, the field has moved toward Self-Supervised Learning (SSL)~\cite{xu2020spatial,liu2023spatio}, endeavoring to distill transferable ST representations. (1) \emph{Contrastive Learning} approaches maximize the mutual information between augmented views of ST graphs. Methods such as ST-SSL~\cite{ji2023spatio} and STGCL~\cite{liu2022contrastive} employ dual-level objectives and diverse data augmentations to enhance representation robustness. (2) \emph{Generative Learning} approaches, inspired by Masked Autoencoders (MAE), learn by reconstructing masked ST segments. For example, STEP~\cite{shao2022pre} utilizes a scalable pretraining framework on long-term historical data to enhance downstream forecasting. While these methods introduce the concept of ``pretraining'' to ST mining, they largely remain confined to \emph{single-domain pretraining} (e.g., pretraining on traffic data to predict traffic). They grapple with the ``Pattern Mismatch'' when transferring across physically distinct domains (e.g., from power grids to traffic networks), as they struggle to disentangle universal temporal laws from domain-specific spatial constraints.

\vspace{0.3em}
\noindent\textbf{ST Foundation Models (STFMs).} 
The recent surge in Foundation Models (FMs) has catalyzed efforts to build universal ST learners capable of \emph{cross-domain generalization}~\cite{liang2025foundation,goodge2025spatio}. Current STFMs generally fall into two categories: (1) \emph{LLM-based Adapters:} Methods like ST-LLM~\cite{liu2024spatial} and UrbanGPT~\cite{li2024urbangpt} leverage the reasoning capacities of Large Language Models by aligning ST data with textual tokens. While promising for few-shot reasoning, they typically suffer from high inference latency and information loss during tokenization. (2) \emph{End-to-End Large ST Models:} Models like UniST~\cite{yuan2024unist} and OpenCity~\cite{li2024opencity} adopt Transformer architectures to jointly model spatial and temporal dependencies on massive datasets. UniST employs learnable prompts to distinguish domains in grid formats, while OpenCity integrates graph transformers for topological flexibility. However, these ``Joint ST Learning'' approaches incur quadratic complexity ($\mathcal{O}(N^2T^2)$) and risk negative transfer due to the conflicting spatial topologies inherent in joint pretraining. Conversely, Time Series FMs (e.g., TimesFM~\cite{das2024decoder}, Chronos~\cite{ansari2024chronos}, Rose~\cite{wangtowards}) achieve efficiency via pure temporal pretraining but lack spatial awareness. \method bridges this gap by proposing a \emph{factorized paradigm}: it first learns universal temporal patterns scalably, then injects lightweight spatial adapters for rapid ST adaptation, achieving versatility and efficiency without the heavy cost of joint ST pretraining.
\begin{figure*}[t!]
    \centering
    \includegraphics[width=\linewidth]{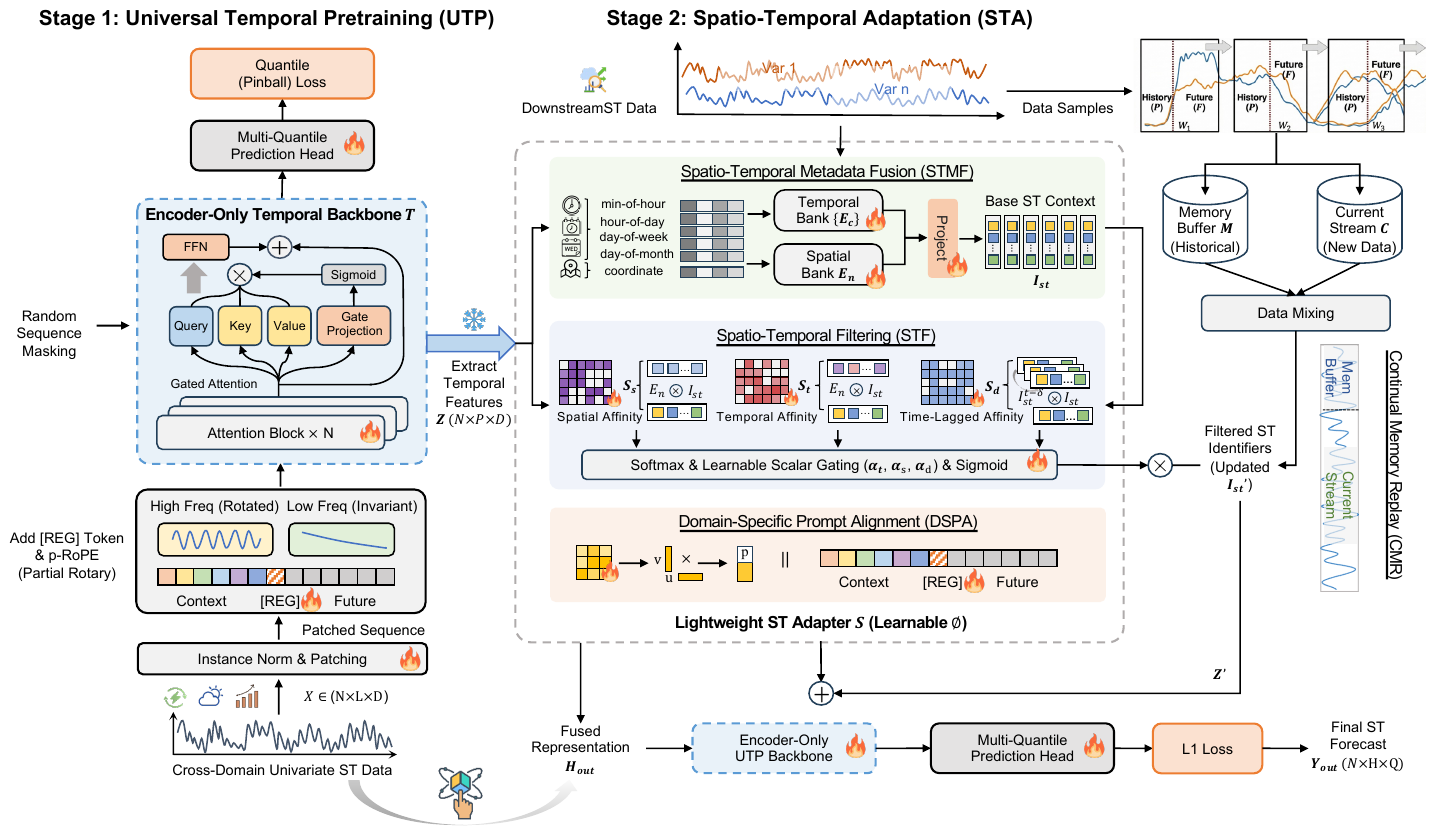}
    \caption{Overview of \method, consisting of two stages: Universal Temporal Pretraining (UTP) for general temporal feature learning (via instance normalization, semantic-preserving positional encoding, gating attention, and quantile prediction) and Spatio-Temporal Adaptation (STA) for domain-specific adaptation (via metadata fusion, filtering, prompt alignment, and continual memory replay), enabling effective cross-domain transfer for ST sequences.}
    \label{fig:framework}
\end{figure*}

\section{Methodology}

\autoref{fig:framework} presents the framework of \method for factorized STFMs. The framework consists of two distinct stages:

\begin{itemize}[leftmargin=*]
    \item \textbf{Stage I (Universal Temporal Pretraining):} Given cross-domain univariate ST data $\mathbf{X} \in \mathbb{R}^{N \times L \times D}$ (where $N$ denotes spatial nodes, $L$ input time steps, and $D$ feature dimension), we first segment each node-wise sequence into $P$ non-overlapping patches. A compact, general-purpose temporal backbone $\mathcal{T}$ is then applied to extract temporal features. During training, we randomly sample variable-length windows to extract context-target pairs, enabling the model to handle sequences of varying lengths. This stage utilizes instance normalization, semantic-preserving positional encoding, gating attention, and quantile prediction to distill universal temporal patterns (e.g., periodicity, trends) from massive data without explicit spatial modeling.

    \item \textbf{Stage II (Spatio-Temporal Adaptation):} For a downstream ST domain, we attach a lightweight adapter $\mathcal{S}$ to the pretrained $\mathcal{T}$. Reusing $\mathcal{T}$ to extract node-wise temporal features $\mathbf{Z} \in \mathbb{R}^{N \times P \times d}$ (where $d$ is the hidden dimension), $\mathcal{S}$—parameterized by $\phi$ with $\|\phi\| \ll \|\Theta_{\mathcal{T}}\|$—refines features via ST metadata fusion, ST filtering, prompt alignment, and memory replay for fine-tuning. The final output $\mathbf{Y}_{\text{out}} = \mathcal{S}(\mathbf{Z}; \mathbf{M}) \in \mathbb{R}^{N \times H \times D}$ yields $H$-step forecasts.
\end{itemize}

\subsection{Stage I: Universal Temporal Pretraining (UTP)}
\label{sec:utp}

The UTP stage aims to distill universal temporal laws from massive, heterogeneous time-series corpora. In contrast to the preliminary version (FactoST-v1 \cite{zhonglearning}), which relied on a rigid Encoder-Decoder architecture and deterministic forecasting, FactoST-v2 introduces a streamlined \textit{Encoder-Only Backbone}.
This redesign specifically targets three limitations of the previous iteration: 1) \textbf{rigidity in variable-length adaptation}, 2) \textbf{semantic degradation caused by positional encoding}, and 3) \textbf{incapability of quantifying predictive uncertainty}. As illustrated in Figure \ref{fig:framework}(a), the new UTP framework comprises the following synergistic components:

\vspace{0.3em}
\noindent\textbf{Normalization and Patching.}
Data from diverse domains exhibit significant scale variations and non-stationarity. To ensure stable cross-domain pretraining, we first adopt instance normalization \cite{kim2021reversible} to standardize the input distribution. Given a univariate input $\mathbf{x} \in \mathbb{R}^{L}$, we center and scale it to zero mean and unit variance, subsequently tokenizing the normalized sequence into non-overlapping patches $\mathbf{Z}_{ctx} \in \mathbb{R}^{N_{ctx} \times D}$:
\begin{equation}
    \mathbf{x}_{norm} = \frac{\mathbf{x} - \mu(\mathbf{x})}{\sigma(\mathbf{x}) + \epsilon},
\end{equation}
where $\mu(\cdot)$ and $\sigma(\cdot)$ denote instance-wise statistics.

\vspace{0.3em}
\noindent\textbf{Random Sequence Masking.} 
The previous Encoder-Decoder architecture \cite{zhonglearning} necessitated fixed input/output horizons during pretraining, restricting the model's ability to adapt to downstream tasks with varying sequence lengths. In this version, we propose a \textit{Randomized Sequence Masking} strategy coupled with a learnable \textit{register token} ($\mathbf{e}_{reg}$). We define a maximum context capacity $L_{max}$ and a maximum prediction horizon $H_{max}$. During pretraining, for each sample, we dynamically sample a mask length $l_{mask}$ from a uniform distribution:
\begin{equation}
    l_{mask} \sim \mathcal{U}(0, L_{max} - L_{min}),
\end{equation}
where $L_{min}$ is the minimum informative context length. We mask the initial $l_{mask}$ patches of the input context to zero. The input sequence $\mathbf{H}^{(0)}$ is then constructed as:
\begin{equation}
    \mathbf{H}^{(0)} = \left[ \text{Mask}(\mathbf{Z}_{ctx}) \parallel \mathbf{e}_{reg} \parallel \mathbf{Z}_{fut} \right] \mathbf{W}_{proj}.
\end{equation}
The $\mathbf{e}_{reg}$ token serves as a semantic pivot, distinguishing the historical context from future placeholders $\mathbf{Z}_{fut}$. By exposing the model to varying effective context lengths $L_{eff} \in [L_{min}, L_{max}]$ within a fixed architecture, the backbone learns a length-agnostic mapping function. This facilitates flexible inference: downstream tasks with shorter horizons are handled via direct truncation, while longer horizons are supported via rolling prediction using the same pretrained head.

\vspace{0.3em}
\noindent\textbf{Semantic-aware Positional Encoding \& Gating.}
Instead of using absolute position embeddings which lack length extrapolation capability, our UTP adopts Partial Rotary Positional Embedding (p-RoPE) \cite{barbero2024round}, which selectively applies rotary operations based on frequency. We decompose the embedding space into high-frequency ($\mathbf{x}_{high}$) and low-frequency ($\mathbf{x}_{low}$) subspaces. The rotary matrix $\mathbf{R}_{\Theta, m}$ is applied exclusively to $\mathbf{x}_{high}$ to capture sequential order, while $\mathbf{x}_{low}$ remains invariant to preserve semantic amplitude:
\begin{equation}
    \text{p-RoPE}(\mathbf{x}, m) = \left[ \mathbf{x}_{high} \otimes \mathbf{R}_{\Theta, m} \parallel \mathbf{x}_{low} \right].
\end{equation}
This design ensures position awareness while maintaining the physical semantics of the signal. Furthermore, we incorporate a \textit{Gated Attention} mechanism \cite{qiu2025gated} to filter noise and address the attention sink problem \cite{xiao2023efficient}. A gating score $\mathbf{G}$ is computed parallel to the query to modulate the attention output:
\begin{equation}
    \mathbf{O} = \text{Attention}(\mathbf{Q}', \mathbf{K}', \mathbf{V}) \odot \sigma(\mathbf{G}),
\end{equation}
where $\mathbf{Q}', \mathbf{K}'$ are position-encoded via p-RoPE.

\vspace{0.3em}
\noindent\textbf{Probabilistic Quantile Forecasting.}
The preliminary version relied on deterministic point prediction, failing to capture the stochastic nature of real-world systems or provide confidence intervals. Here, we adopt a \textit{Multi-Quantile Prediction Head}. Instead of a single value, the model predicts the conditional distribution at multiple quantile levels $q \in \mathcal{Q}$ (e.g., $\{0.1, 0.5, 0.9\}$) \cite{wen2017multi}, optimized via the Pinball Loss:
\begin{equation}
    \mathcal{L}_{UTP} = \frac{1}{|\mathcal{Q}|} \sum_{q \in \mathcal{Q}} \max\left( (q-1)(\mathbf{y} - \hat{\mathbf{y}}_q), \; q(\mathbf{y} - \hat{\mathbf{y}}_q) \right).
\end{equation}
This probabilistic objective compels the backbone to learn the full distributional properties of temporal dynamics, significantly enhancing its utility for high-stakes decision-making.

\subsection{Stage II: Spatio-Temporal Adaptation}
To adapt the pretrained temporal backbone to ST scenarios, we introduce three lightweight modules that incur minimal parameter overhead while effectively capturing ST dependencies.

\vspace{0.3em}
\noindent\textbf{Spatio-Temporal Metadata Fusion (STMF).}
This module provides fundamental ST context by injecting explicit, learnable identifiers into the temporal backbone. Given ST input $\mathbf{X} \in \mathbb{R}^{N \times L \times D}$, we first obtain temporal representations $\mathbf{Z} \in \mathbb{R}^{N \times P \times d}$ via the pretrained backbone. We then define: (1) a spatial bank $\mathbf{E}_n \in \mathbb{R}^{N \times d}$ encoding unique node characteristics; (2) a calendar-aware temporal embedding bank $\{\mathbf{E}_c\}_{c \in \mathcal{S}}$ for multi-scale cycles—each $c$ (e.g., minute-of-hour, day-of-week) maps to an embedding table $\mathbf{E}_c \in \mathbb{R}^{K_c \times d}$ capturing distinct cyclic granularities. For each node-patch pair $(i,\tau)$, we retrieve the corresponding spatial embedding $\mathbf{E}_n^i$ and temporal embeddings based on discrete time features, which are concatenated and projected to the hidden dimension:
\begin{equation}
\mathbf{I}_{\text{st}}(i,\tau) = \mathbf{W}_{\text{meta}} \left[ \mathbf{E}_n^i \big\|\big\|_{c\in\mathcal{S}} \mathbf{E}_c^{\phi_c(\tau)}\right],
\end{equation}
where $\|$ is concatenation and $\mathbf{W}_{\text{meta}}$ is a linear projection. This creates the base ST context $\mathbf{I}_{\text{st}}$ that endows the graph-agnostic backbone with explicit spatial and temporal awareness.


\vspace{0.3em}
\noindent\textbf{Spatio-Temporal Filtering (STF).}
While STMF provides raw identifiers, STF adapts to scenario-dependent cue relevance—e.g., local spatial context for incidents versus global temporal patterns for rush hours—by dynamically reweighting these identifiers. Given the output $\mathbf{I}_{\text{st}}$ from STMF, and retrieving corresponding spatial $\mathbf{E}_n$ and temporal embeddings $\mathbf{E}_t$ (aggregated from STMF's bank), we compute three affinities:

\begin{itemize}[leftmargin=*]
    \item \textbf{Spatial Affinity ($\mathbf{S}_s$):} Measures the compatibility between $\mathbf{I}_{\text{st}}$ and its spatial component $\mathbf{E}_n$ via the dot product: $\mathbf{S}_s = \langle \mathbf{I}_{\text{st}},\, \mathbf{E}_n \rangle \in \mathbb{R}^{N \times P}$. Higher values indicate stronger spatial relevance for each (node, patch) pair. This dynamic measurement reduces rigid reliance on fixed spatial identifiers by adapting to context-specific spatial relevance.

    \item \textbf{Temporal Affinity ($\mathbf{S}_t$):} Quantifies the alignment between $\mathbf{I}_{\text{st}}$ and its temporal component $\mathbf{E}_t$ via the pointwise dot product: $\mathbf{S}_t = \langle \mathbf{I}_{\text{st}},\, \mathbf{E}_t \rangle \in \mathbb{R}^{N \times P}$. It captures how well the current ST context matches dominant temporal patterns, while filtering out transient or irrelevant temporal noise.

    \item \textbf{Time-Lagged Affinity ($\mathbf{S}_d$):} Models asynchronous causal effects (e.g., delayed influence of upstream nodes). To capture global interactions efficiently, we introduce a set of learnable latent prototypes $\mathbf{P}_{lat} \in \mathbb{R}^{M \times d}$. We aggregate historical identifiers $\mathbf{I}_{\text{st}}^{(t-\delta)}$ by projecting them onto these latent prototypes and computing the similarity with current $\mathbf{I}_{\text{st}}$, weighted by learnable lag coefficients $\gamma^{(\delta)}$: $\mathbf{S}_d = \sum_{\delta=1}^{\Delta} \gamma^{(\delta)} \cdot \langle \mathbf{I}_{\text{st}},\, \text{Agg}_{\delta}(\mathbf{I}_{\text{st}}^{(t-\delta)}; \mathbf{P}_{lat}) \rangle$, where $\text{Agg}_{\delta}$ denotes the latent-projection-based aggregation.
\end{itemize}

Distinct from complex projection schemes, we employ a lightweight, learnable scalar weighting mechanism to fuse these affinities. We define three learnable scalar parameters $w_s, w_t, w_d$ and compute normalized fusion weights via Softmax: $\boldsymbol{\alpha} = \text{softmax}([w_s, w_t, w_d])$. The final gate is generated by a weighted summation passed through a Sigmoid activation:
\begin{equation}
\mathbf{I}_{\text{st}} = \sigma\left( \alpha_s \mathbf{S}_s + \alpha_t \mathbf{S}_t + \alpha_d \mathbf{S}_d \right) \odot \mathbf{I}_{\text{st}}
\end{equation}
The final representation after filtering is the summation of the refined temporal features and the filtered ST identifiers: $\mathbf{H}_{\text{fused}} = \mathbf{Z} + \mathbf{I}_{\text{st}}$. Instead of feeding this directly into the prediction head, we first align it with the target domain distribution via the subsequent prompt module.

\vspace{0.3em}
\noindent\textbf{Domain-Specific Prompt Alignment (DSPA).}
While STMF handles local context, there remains a global distribution shift between the pretraining corpora and the downstream domain. To align these distributions without fine-tuning the heavy backbone, we employ a prompt-based adaptation strategy. We introduce a set of global learnable tokens $\mathbf{P} \in \mathbb{R}^{K \times d}$. To ensure parameter efficiency, we use a low-rank \cite{hu2022lora} factorization:
\begin{equation}
\mathbf{P} = \mathbf{U}\mathbf{V}^\top, \quad \mathbf{U} \in \mathbb{R}^{K \times r}, \mathbf{V} \in \mathbb{R}^{d \times r},
\end{equation}
where $r \ll d$ (rank) and $K$ is the number of tokens. These prompt tokens serve as learnable global contexts capturing implicit distributional semantics of the downstream data, and are prepended to $\mathbf{H}_{\text{fused}}$ as an optimizable prefix \cite{li2021prefix}:
\begin{equation}
\mathbf{H}_{out} = [\mathbf{P} \parallel \mathbf{H}_{\text{fused}}].
\end{equation}
Finally, the prompted sequence $\mathbf{H}_{out}$ is fed into the UTP backbone encoder to capture high-order dependencies via rotary attention. The final representation is projected by the predictor to enable flexible objective choices: for probabilistic tasks, we optimize the Pinball loss for uncertainty quantification; for deterministic benchmarks, we adopt L1 loss on the median quantile ($\tau=0.5$) to ensure fair alignment with prevailing point-forecasting baselines.

\vspace{0.3em}
\noindent\textbf{Continual Memory Replay (CMR).}
To mitigate catastrophic forgetting during few-shot adaptation, we implement dynamic data mixing, combining current data with historical samples. First, we establish a memory buffer. Given a stream of training samples $\{\mathbf{X}_k\}_{k=1}^{K_{total}}$, we partition the dataset into:
\begin{equation}
    \underbrace{\mathcal{M} = \{\mathbf{X}_k\}_{k=1}^{K_m}}_{\text{memory buffer}} 
    , \quad 
    \underbrace{\mathcal{C} = \{\mathbf{X}_k\}_{k=K_m+1}^{K_{total}}}_{\text{current stream}}
\end{equation}
where $K_m = \lfloor s \cdot K_{total} \rfloor$ ($s$ is memory size). 

Specifically, the memory buffer $\mathcal{M}$ retains key temporal patterns from initial learning, ensuring stability across domain shifts. Each mini-batch $\mathcal{B}$ is constructed by mixing samples from both current stream and memory buffer: $\mathcal{B} = \{\mathbf{X}_i\}_{i\in\mathcal{I}_c \backslash \mathcal{R}} \cup \{\mathbf{X}_j\}_{j\in\mathcal{I}_m[:\lvert\mathcal{R}\rvert]}$, where $\mathcal{I}_c$ and $\mathcal{I}_m$ denote shuffled indices of $\mathcal{C}$ and $\mathcal{M}$, respectively, and $\mathcal{R} \subset \mathcal{I}_c$ has size $\lfloor r \cdot |\mathcal{B}| \rfloor$. This mechanism effectively preserves implicit historical knowledge during domain adaptation. 

\begin{algorithm}[h]
\caption{Pseudocode of \method. (\textcolor{utpColor}{\textbf{Blue}}: UTP Upgrades; \textcolor{staColor}{\textbf{Red}}: STA Upgrades)}
\label{alg:factost_procedure}
\begin{algorithmic}[1]
\Require 
    Pretraining data $\mathcal{D}_{pre}$; Downstream data $\mathcal{D}_{tgt}$; 
    Backbone $\mathcal{T}$ with parameters $\theta$; Adapter $\mathcal{S}$ with parameters $\phi$ (including STMF, STF, and \textcolor{staColor}{DSPA Prompts $\mathbf{P}$});
    Memory size $S_{mem}$.
\Ensure Optimized parameters $\theta, \phi$ for target domain.

\Statex \textbf{// Stage I: Universal Temporal Pretraining (UTP)}
\State Initialize \textcolor{utpColor}{Encoder-Only} backbone $\theta$ randomly.
\While{not converged}
    \State Sample batch $\mathbf{X} \sim \mathcal{D}_{pre}$.
    \State Apply normalization and \textcolor{utpColor}{random masking strategy}.
    \State Forward pass: $\hat{\mathbf{Y}} = \mathcal{T}(\mathbf{X}; \theta)$ generating \textcolor{utpColor}{probabilistic quantiles}.
    \State Compute \textcolor{utpColor}{Quantile Loss: $\mathcal{L}_{UTP} = \mathcal{L}_{\text{pinball}}(\hat{\mathbf{Y}}, \mathbf{Y})$}.
    \State Update $\theta \leftarrow \theta - \eta \nabla_\theta \mathcal{L}_{UTP}$.
\EndWhile
\State \textbf{Output:} Pretrained Universal Backbone $\mathcal{T}(\theta_{pre})$.

\Statex \textbf{// Stage II: Spatio-Temporal Adaptation (STA)}
\State Initialize Adapter $\phi$ (including \textcolor{staColor}{prompts $\mathbf{P}$}); Initialize Memory Buffer $\mathcal{M} \leftarrow \emptyset$.
\State Load pretrained weights: $\theta \leftarrow \theta_{pre}$.
\State Define learnable parameters: $\Theta = \{\theta, \phi\}$.

\Statex \textit{\quad $\rhd$ Continual Memory Replay (CMR)}
\For{epoch $e = 1 \to E_{adapt}$}
    \For{current stream batch $\mathcal{B}_{curr} \sim \mathcal{D}_{tgt}$}
        \State Sample memory batch $\mathcal{B}_{mem} \sim \mathcal{M}$.
        \State Mix data: $\mathcal{B}_{mix} = \text{Mix}(\mathcal{B}_{curr}, \mathcal{B}_{mem})$.
        \State \textbf{Forward:} 
        \State \quad $\mathbf{Z} \leftarrow \mathcal{T}_{\text{emb}}(\mathbf{X}_{mix}; \theta)$.
        \State \quad $\mathbf{I}_{st} \leftarrow \text{STMF}(\mathbf{Z}, \mathbf{M}_{meta})$.
        \State \quad $\mathbf{I}'_{st} \leftarrow \text{STF}(\mathbf{I}_{st})$.
        \State \quad $\mathbf{H}_{fused} \leftarrow \mathbf{Z} + \mathbf{I}'_{st}$.
        \State \quad \textcolor{staColor}{$\mathbf{H}_{in} \leftarrow \text{Concat}(\mathbf{P}, \mathbf{H}_{fused})$}. \Comment{\textit{Inject Prompts}}
        \State \quad $\mathbf{H}_{enc} \leftarrow \mathcal{T}_{\text{enc}}(\mathbf{H}_{in}; \theta)$.
        \State \quad Forecast $\hat{y} = \mathcal{T}_{\text{head}}(\mathbf{H}_{enc}; \theta)$.
        \State \textbf{Backward:} 
        \State \quad Compute Task Loss \textcolor{staColor}{$\mathcal{L}_{l1}(\hat{y}, y)$}. \Comment{\textit{Pure Forecasting}}
        \State \quad Update $\Theta \leftarrow \Theta - \eta \nabla_{\Theta} \mathcal{L}_{l1}$.
        \State Update Memory $\mathcal{M}$ with samples from $\mathcal{B}_{curr}$.
    \EndFor
\EndFor

\State \Return Adapted Model with parameters $\theta, \phi$.
\end{algorithmic}
\end{algorithm}

\subsection{Theoretical Analysis: Scalability and Generalization}
\label{sec:theoretical_analysis}

To rigorously justify the superiority of the factorized paradigm, we provide a theoretical analysis from two perspectives: \textit{asymptotic complexity} and \textit{generalization bounds}.

\vspace{0.3em}
\noindent\textbf{Computational Complexity}
\label{sec:complexity}
A critical bottleneck in existing STFMs is the scalability w.r.t. spatial nodes $N$.

\begin{itemize}[leftmargin=*]
    \item \textbf{Joint Modeling Paradigm.} Standard STFMs flatten spatial and temporal dimensions into a sequence $S = N \times P$. Global self-attention over this sequence incurs quadratic costs:
    \begin{equation}
        \mathcal{O}_{\text{Joint}} = \mathcal{O}((NP)^2 D) \approx \mathcal{O}(N^2 P^2 D).
    \end{equation}
    This $\mathcal{O}(N^2)$ dependence creates a severe scalability wall, causing memory explosion on large-scale graphs and exacerbating the over-squashing problem~\cite{marisca2025over}.

    \item \textbf{Our Factorized STFM.} As shown in~\autoref{tab:complexity_comparison}, our factorized modeling efficiently avoids the heavy computations of existing STFMs. \textbf{In UTP}: The backbone processes node-wise series independently, yielding linear complexity w.r.t. $N$: $\mathcal{O}_{\text{UTP}} = \mathcal{O}(N P^2 D)$. Since $P \ll N$ in large-scale ST mining, removing the $N^2$ term drastically reduces overhead. \textbf{In STA}: Adapter modules achieve linear scalability: (1) \textit{STMF}: linear projections ($\mathcal{O}(N P D)$); (2) \textit{STF}: pointwise dot products ($\mathcal{O}(NPD)$) and latent global interactions via $M \ll N$ prototypes ($\mathcal{O}(\Delta N P M D)$). The added DSPA involves only constant overhead per sequence. Consequently, the total adaptation complexity scales linearly ($\mathcal{O}(N)$).

\end{itemize}

\begin{table}[h]
    \centering
    \small
    \caption{Asymptotic complexity comparison per layer. \method achieves linear scalability w.r.t. spatial nodes $N$.}
    \label{tab:complexity_comparison}
    \scalebox{0.80}{
    \begin{tabular}{l@{\hspace{1em}}c@{\hspace{1em}}c}
        \toprule
        \textbf{Model Paradigm} & \textbf{Complexity} & \textbf{Scalability w.r.t. $N$} \\
        \midrule
        Joint ST (e.g., UniST) & $\mathcal{O}(N^2 P^2 D)$ & Quadratic $\mathcal{O}(N^2)$ \\
        Pure Temporal (e.g., TimesFM) & $\mathcal{O}(N P^2 D)$ & Linear $\mathcal{O}(N)$ \\
        \textbf{\method (Ours)} & $\mathbf{\mathcal{O}(N P^2 D + N M P D)}$ & \textbf{Linear $\mathcal{O}(N)$} \\
        \bottomrule
    \end{tabular}}
\end{table}

\vspace{0.3em}
\noindent\textbf{Generalization Bounds.}
\label{sec:generalization}
We further investigate why factorization improves downstream accuracy. Let $\mathcal{H}$ denote the hypothesis space and $\mathfrak{C}(\mathcal{H})$ its complexity measure (e.g., Rademacher complexity \cite{bartlett2002rademacher}). By standard learning theory~\cite{mohri2018foundations}, for any $h \in \mathcal{H}$, the expected risk $R(h)$ is bounded with probability $1-\delta$ by the empirical risk $\hat{R}(h)$ plus a complexity term:
\begin{equation}
    R(h) \leq \hat{R}(h) + \mathcal{O}\left(\sqrt{\frac{\mathfrak{C}(\mathcal{H})}{|\mathcal{D}|}}\right),
\end{equation}
where $|\mathcal{D}|$ represents the sample size.

In downstream adaptation (ranging from zero/few-shot to full-shot), the generalization error is governed by the trade-off between fitting the data $\hat{R}(h)$ and controlling model complexity $\mathfrak{C}(\mathcal{H})$. We compare the two paradigms as follows:

\begin{enumerate}
    \item \textbf{Bound Looseness in Joint Modeling.} Joint STFMs learn cross-domain coupled mappings. Their hypothesis space $\mathcal{H}_{\text{joint}}$ must cover vast domain-specific spatial topology variability, leading to an extremely large $\mathfrak{C}(\mathcal{H}_{\text{joint}})$. In zero/few-shot scenarios ($|\mathcal{D}| \to 0$), this large complexity yields a trivial bound, indicating high risk of negative transfer. Even in full-shot settings, heterogeneous topology structural conflicts often prevent $\hat{R}(h)$ from reaching global optimum, causing optimization difficulties.

    \item \textbf{Tighter Bounds via Factorization.} \method decomposes the final hypothesis into $h = h_{\text{adapt}} \circ h_{\text{time}}$. This splits the learning bound into two tractable components: 
    1) \textit{Universal Backbone Robustness}: $h_{\text{time}}$ captures invariant temporal patterns from massive pretraining ($|\mathcal{D}_{\text{pre}}| \gg \mathfrak{C}(\mathcal{H}_{\text{time}})$), ensuring negligible generalization error for the temporal component. 
    2) \textit{Restricted Adaptation Space}: Our adapter is structurally constrained with low complexity, i.e., $\mathfrak{C}(\mathcal{H}_{\text{adapt}}) \ll \mathfrak{C}(\mathcal{H}_{\text{joint}})$. 
    Consequently, for any target dataset size $|\mathcal{D}_{tgt}|$, the generalization bound depends primarily on the light adapter rather than the monolithic model capacity, yielding a significantly tighter bound:
    \begin{equation}
        \underbrace{\sqrt{\frac{\mathfrak{C}(\mathcal{H}_{adapt})}{|\mathcal{D}_{tgt}|}}}_{\text{\method (Tight Bound)}} \ll \underbrace{\sqrt{\frac{\mathfrak{C}(\mathcal{H}_{joint})}{|\mathcal{D}_{tgt}|}}}_{\text{Joint ST (Loose Bound)}}.
    \end{equation}
\end{enumerate}

This inequality theoretically substantiates our empirical observation: decoupling universal temporal patterns from spatial is a prerequisite for robust generalization across the full spectrum of data regimes in heterogeneous ST domains.

\vspace{0.3em}
\noindent\textbf{Methodological Evolution beyond V1.} 
Compared to the v1 version \cite{zhonglearning}, this journal iteration (\method) strategically simplifies the architecture to maximize transfer. The procedural details are outlined in Algorithm~\ref{alg:factost_procedure}. In the \textbf{UTP stage}, we shift from a rigid encoder-decoder to a \textcolor{utpColor}{minimalist \textit{encoder-only} design}; enabled by \textcolor{utpColor}{random sequence masking} and \textcolor{utpColor}{probabilistic forecasting}, the backbone internalizes diverse temporal dynamics. 
Critically, in the \textbf{STA stage}, we replace complex hierarchical alignment with a \textcolor{staColor}{streamlined DSPA via Prompting}. 
This modification eliminates task redundancy (focusing solely on \textcolor{staColor}{pure forecasting}) and aligns the backbone via \textcolor{staColor}{learnable prompt tokens}, achieving superior adaptation.
\section{Experiments}
In our experiments, we address the following research questions (RQs) to comprehensively evaluate \method:

\begin{itemize}[leftmargin=*]
    \item \textbf{RQ1 (Overall Performance)}: Can \method establish state-of-the-art (SOTA) performance across diverse forecasting scenarios, including \textit{few-shot}, \textit{full-shot}, and \textit{zero-shot} settings, compared to existing STFMs and domain-specific expert models? $\Rightarrow$ \textbf{Sec. 4.2}, \textbf{Sec. 4.3} \& \textbf{Sec. 4.4}.
    
    \item \textbf{RQ2 (Ablation Study)}: How do the key architectural designs in the two-stage factorized framework—spanning the \textit{UTP} mechanisms and the \textit{STA} modules—contribute to the final prediction accuracy? $\Rightarrow$ \textbf{Sec. 4.5}.

    \item \textbf{RQ3 (Scaling Properties)}: Does \method exhibit  scaling laws with respect to \textit{model parameter size}, and how data-efficient is it during downstream fine-tuning? $\Rightarrow$ \textbf{Sec. 4.6}.

    \item \textbf{RQ4 (In-depth Analysis)}: Can \method yield interpretable adaptation insights in STF module, maintain robustness across \textit{temporal granularities}, and demonstrate \textit{architectural generality} across backbones? $\Rightarrow$ \textbf{Sec. 4.7}.

    \item \textbf{RQ5 (Efficiency)}: How does the \textit{computational efficiency} \method compare to traditional joint spatio-temporal modeling paradigms? $\Rightarrow$ \textbf{Sec. 4.8}.
\end{itemize}

\subsection{Experiment Setting}

\vspace{0.3em}
\noindent\textbf{Datasets.} 
To build a universal temporal backbone, we expand the pretraining corpus beyond conventional ST studies. Per \autoref{tab:pretraining_stats}, our curated repository spans eight diverse domains—physical systems (Energy, Weather, Transport, Healthcare), human activities (Economics, Web), and forecasting competitions (e.g., M4, M5). Notably, we incorporate KernelSynth \cite{ansari2024chronos}, a massive synthetic dataset generating kernel-based patterns, to bridge the gap between finite real-world data and infinite temporal dynamics. Using BLAST \cite{shao2025blast} for sampling, we yield a training scale of over \textbf{11B} time points (see Table \ref{tab:pretraining_stats}). During UTP, we treat these as \textbf{independent node-wise series} to enforce task-agnostic temporal representation learning. For downstream adaptation, we adopt eight standard ST benchmarks: traffic flow (PEMS03/04/07/08), road speed (PEMS-BAY), electricity (ECL), transformer temperature (ETTh2), and climate (Weather). These datasets cover resolutions from 5-min to daily and topologies of 21–883 nodes, forming a rigorous testbed for \method's cross-domain transferability. All evaluation datasets are excluded from the pretraining corpus to avoid data leakage and ensure fair zero-shot evaluation.

\begin{table}[h!]
\centering
\scriptsize
\setlength{\tabcolsep}{2pt}
\renewcommand{\arraystretch}{0.9} 

\caption{Summary of Pretraining and Evaluation Datasets.}
\label{tab:pretrain_eval_datasets}

\begin{tabular*}{\linewidth}{@{\extracolsep{\fill}} l | c | c | c | c | c }
\toprule
\textbf{Type} & \textbf{Category} & \textbf{Dataset} & \textbf{Freq.} & \textbf{\# Feat.} & \textbf{\# Time Pts.} \\
\midrule

\multirow{22}{*}{\rotatebox[origin=c]{90}{\textbf{Pretraining}}} 
& \multirow{3}{*}{Forecasting} & M1/M3/M4 Comp. & Mixed & -- & 30.0M \\
& & M5 (Walmart) & Daily & -- & 59.2M \\
& & Tourism (Large) & Mixed & -- & 0.5M \\
\cmidrule{2-6}

& \multirow{4}{*}{Energy \& Power} & London Smart Meters & 30min & -- & $\sim$100M \\
& & Solar Power & 10min & -- & 7.2M \\
& & Wind Power & Day/Hr & -- & 17.8M \\
& & Aus. Elec. Demand & 30min & -- & 0.3M \\
\cmidrule{2-6}

& \multirow{3}{*}{Transport} & NYC Taxi / Uber & 30min & -- & 4.6M \\
& & Melb. Pedestrian & Hourly & -- & 3.0M \\
& & Saugeenday & Daily & -- & 23.7k \\
\cmidrule{2-6}

& \multirow{3}{*}{Weather \& Nature} & Global Weather & Daily & -- & 10.5M \\
& & ERA5 Subset (Grid) & Hourly & -- & 10.0M \\
& & Rainfall & Month & -- & 1.2M \\
\cmidrule{2-6}

& \multirow{2}{*}{Web \& Cloud} & Wiki2000 / Wiki & Daily & -- & 116M \\
& & Server Usage (Ali/Goog) & 5min & -- & $\sim$100M \\
\cmidrule{2-6}

& \multirow{4}{*}{Econ. \& Finance} & Exchange Rate & Daily & -- & 46.0k \\
& & Dominick's (Retail) & Weekly & -- & 45.0M \\
& & NN5 (ATM) & Daily & -- & 62.0k \\
& & Fred-MD & Month & -- & 0.1M \\
\cmidrule{2-6}

& \multirow{3}{*}{General \& Synth.} & UCR Archive & Mixed & -- & $\sim$50M \\
& & Car Parts & Month & -- & 0.1M \\
& & KernelSynth & Any & -- & \textbf{11.0B} \\

\midrule
\midrule

\multirow{10}{*}{\rotatebox[origin=c]{90}{\textbf{Evaluation}}} 
& \multirow{4}{*}{Traffic Flow} & PEMS03 & 5min & 358 & 26.2k \\
& & PEMS04 & 5min & 307 & 17.0k \\
& & PEMS07 & 5min & 883 & 28.2k \\
& & PEMS08 & 5min & 170 & 17.9k \\
\cmidrule{2-6}
& \multirow{2}{*}{Traffic Speed} & PEMS-BAY & 5min & 325 & 52.1k \\
& & METR-LA & 5min & 207 & 34.2k \\
\cmidrule{2-6}
& Power & ETTh2 & 1h & 7 & 14.4k \\
\cmidrule{2-6}
& Load & Electricity & 1h & 321 & 26.3k \\
\cmidrule{2-6}
& Meteorology & Weather & 10min & 21 & 52.7k \\

\bottomrule
\end{tabular*}
\end{table}
\begin{table}[h!]
  \centering
  \caption{Pretraining corpora and scale of STFMs and TSFMs.}
  \setlength{\tabcolsep}{3pt}
  \renewcommand{\arraystretch}{1.1}
  \resizebox{\linewidth}{!}{
  \begin{tabular}{l|l|l|c}
    \toprule
    \textbf{Category} & \textbf{Model} & \textbf{Pretraining Coverage} & \textbf{Scale (Obs.)} \\
    \midrule
    
    \multirow{2}{*}{\textbf{Ours}} 
    & \textit{FactoST-v1 (Conf.)} & Real-world Monash (6 Domains) & $\sim$13 M \\
    & \textbf{FactoST-v2 (Jour.)} & \textbf{8 Domains (Real + KernelSynth)} & \textbf{$\sim$11 B} \\
    \midrule
    
    \multirow{2}{*}{\shortstack[l]{\textbf{Other}\\\textbf{STFMs}}} 
    & OpenCity~\cite{li2024opencity} & Traffic Graphs (10k regions) & 151 M \\
    & UniST~\cite{yuan2024unist} & 6 Grid Datasets (4 Cities) & $\sim$100 M$^*$ \\
    \midrule
    
    \multirow{2}{*}{\shortstack[l]{\textbf{General}\\\textbf{TSFMs}}} 
    & TimesFM~\cite{das2024decoder} & General Time Series (Real+Synth) & $\sim$100 B \\
    & Moirai~\cite{woo2024unified} & LOTSA (Public Datasets) & $\sim$27 B \\
    \bottomrule
  \end{tabular}}
  \label{tab:pretraining_stats}
  \begin{flushleft}
    \vspace{-0.5em}
    \scriptsize{$^*$ Estimated based on standard dataset dimensions (e.g., TaxiBJ/NYC).}
  \end{flushleft}
\end{table}

\begin{table}[h!]
  \centering
  \caption{Hyper-parameter specifications for different model scales.}
  \setlength{\tabcolsep}{8pt}
  \renewcommand{\arraystretch}{1}
  \resizebox{\linewidth}{!}{
  \begin{tabular}{l|c|cccc}
    \toprule
    \textbf{Variant} & \textbf{Parameters} & \boldmath{$d_{\text{model}}$} & \boldmath{$d_{\text{ff}}$} & \boldmath{$n_{\text{layers}}$} & \boldmath{$n_{\text{heads}}$} \\
    \midrule
    Minuscule & 2,510,192  & 192 & 768  & 3 & 3 \\
    \textbf{Tiny (Default)}      & 4,378,928  & 256 & 1024 & 3 & 4 \\
    Small      & 12,172,336 & 384 & 1536 & 4 & 6 \\
    Base     & 30,387,632 & 512 & 2048 & 6 & 8 \\
    \bottomrule
  \end{tabular}}
  \label{tab:model_variants}
\end{table}

\vspace{0.3em}
\noindent\textbf{Baselines.} We compare \method with 16 competitive models across four categories, covering both pretrained foundation models and task-specific architectures trained from scratch:

\begin{itemize}[leftmargin=*]
    \item \textbf{Spatio-Temporal Foundation Models (STFMs)}: Includes \textit{UrbanDiT}~\cite{yuan2025urbandit} (diffusion-Transformer for multi-task urban ST prediction), \textit{OpenCity}~\cite{li2024opencity} (STGNN-based pretraining with lightweight ST adapters for downstream task), and \textit{UniST}~\cite{yuan2024unist} (grid-based STFM with prompt learning).

    \item \textbf{Time Series Foundation Models (TSFMs)}: Includes \textit{Rose}~\cite{wangtowards} (lightweight encoder-decoder with frequency decomposition and register tokens), \textit{TimesFM}~\cite{das2024decoder} (decoder-only univariate forecaster for arbitrary horizon point prediction), and \textit{Moirai}~\cite{woo2024unified} (encoder-based universal model for multivariate, multi-frequency probabilistic forecasting).

    \item \textbf{Spatio-Temporal Expert Models (STEMs)}: Includes \textit{BigST}~\cite{han2024bigst} (linear STGNN scalable for large graph networks), \textit{STAEformer}~\cite{liu2023spatio} (Transformer with adaptive ST embeddings), \textit{STID}~\cite{shao2022spatial} (ST identity modeling), \textit{D2STGNN}~\cite{shao2022decoupled} (decoupled spatial and temporal modeling with dynamic graph), and \textit{GWNet}~\cite{wu2019graph} (gated temporal convolutions with adaptive graph construction).

    \item \textbf{Time Series Expert Models (TSEMs)}: Includes \textit{TimeMixer}~\cite{wang2024timemixer} (multiscale time mixed modeling), \textit{PatchTST}~\cite{nie2022time} (patch-based tokenizer), \textit{DLinear}~\cite{zeng2023transformers} (trend-residual decomposed linear model), \textit{FeDformer}~\cite{zhou2022fedformer} (fourier-enhanced decomposed learning), and \textit{Informer}~\cite{zhou2021informer} (sparse attention with generative decoding).

\end{itemize}

\clearpage

\begin{table*}[t!]
  \centering
  \caption{Few-shot Forecasting Comparison. \boldres{Red}=best, \secondres{Blue}=second best. Values: MAE/RMSE. OOM=out-of-memory (NVIDIA A800-80G).}
  \label{tab:few_shot_forecasting}
  \setlength{\tabcolsep}{2pt}
  \footnotesize
  \renewcommand{\arraystretch}{0.8}
  \scalebox{0.95}{
    \begin{tabular}{l|cccccccccc}
      \toprule
      \textbf{Type} & \textbf{Model} 
        & \textbf{PEMS-03} & \textbf{PEMS-04} & \textbf{PEMS-07} & \textbf{PEMS-08} 
        & \textbf{PEMS-Bay} & \textbf{ETTh2} & \textbf{Electricity} & \textbf{Weather} & \textbf{METR-LA} \\
      \midrule
      \multicolumn{11}{c}{\textit{Short-term Forecasting ($12\rightarrow12$)}} \\
      \midrule
      
      \multirow{5}{*}{\shortstack{\textbf{TSEMs}}} 
        & Informer & 23.24 / 37.98& 29.81 / 45.59& 37.55 / 62.55& 31.69 / 51.53& 2.96 / 6.23& 2.125 / 2.898& 1.598 / 15.649& 0.958 / 1.783& 4.93 / 9.20 \\
        & FeDFormer & 19.66 / 30.15& 25.22 / 38.46& 29.41 / 43.96& 20.49 / 31.84& 2.58 / 5.37& 0.788 / 1.266& 0.755 / 4.742& 0.432 / 1.048& 5.31 / 10.55 \\
        & DLinear & 21.94 / 35.30& 28.37 / 44.57& 31.89 / 49.65& 23.10 / 36.35& 2.21 / 5.20& 1.885 / 2.946& 1.282 / 8.837& 0.383 / 1.046& 4.57 / 9.82 \\
        & PatchTST & 21.97 / 35.59& 28.11 / 44.13& 31.19 / 48.91& 22.42 / 35.64& 2.15 / 5.23& 0.721 / 1.211& 0.840 / 5.097& 0.296 / 1.074& 4.34 / 9.75 \\
        & TimeMixer & 21.41 / 33.57& 27.37 / 42.16& 30.31 / 46.36& 22.05 / 34.09& 2.11 / 4.93& 0.803 / 1.228& 0.767 / 4.324& 0.311 / 0.967& 4.23 / 9.20 \\
      \midrule
      
      \multirow{5}{*}{\shortstack{\textbf{STEMs}}}
        & GWNet & \secondres{17.25} / \secondres{27.79}& \secondres{23.27} / \boldres{35.62}& 26.51 / 41.08& \secondres{18.47} / \secondres{29.04}& \secondres{1.91} / \secondres{4.46}& 0.884 / 1.396& 0.772 / 6.161& 0.342 / 0.901& \secondres{3.93} / 8.48 \\
        & D2STGNN & 18.55 / 29.21& 24.86 / 38.43& \secondres{25.51} / \secondres{39.81}& 19.55 / 30.51& 1.99 / 4.72& 0.916 / 1.433& 0.686 / 4.535& 0.587 / 1.269& 4.00 / \secondres{8.03} \\
        & STID & 22.93 / 34.10& 26.72 / 40.31& 31.64 / 46.72& 23.17 / 34.09& 2.00 / 4.57& 0.756 / 1.224& 0.575 / 1.085& 0.330 / 0.920& 4.00 / 8.20 \\
        & STAEformer & 30.79 / 47.67& 48.23 / 68.46& 33.50 / 51.43& 36.15 / 51.05& 2.01 / 4.62& 1.208 / 1.673& 0.858 / 8.289& 0.575 / 1.085& 4.61 / 8.91 \\
        & BigST & 18.41 / 28.45& 23.97 / 36.88& 25.72 / \boldres{39.72}& 19.40 / 29.96& \secondres{1.91} / 4.26& 0.740 / 1.214& 0.638 / 4.545& 0.375 / 0.951& \boldres{3.72} / \boldres{7.19} \\
      \midrule
      
      \multirow{3}{*}{\shortstack{\textbf{TSFMs}}}
        & TimesFM & 21.99 / 35.31& 27.84 / 43.15& 32.61 / 50.20& 22.06 / 33.87& 2.25 / 5.49& 0.284 / 0.410& 0.529 / 0.801& 0.138 / 0.323& 5.56 / 12.87 \\
        & Moirai & 21.40 / 32.38& 33.73 / 54.09& 35.69 / 51.36& 38.01 / 53.05& 2.26 / 5.49& \boldres{0.135} / \boldres{0.307}& 0.837 / 1.036& 0.184 / 0.432& 4.95 / 12.75 \\
        & Rose & 23.00 / 36.21& 30.85 / 46.17& 35.81 / 53.73& 24.79 / 37.44& 2.19 / 5.14& 0.277 / 0.428& 0.536 / 0.772& 0.093 / 0.284& 4.84 / 12.53 \\
      \midrule
      
      \multirow{3}{*}{\shortstack{\textbf{STFMs}}}
        & UniST & 40.39 / 53.44& 42.76 / 59.07& 40.77 / 54.86& 35.70 / 46.74& 5.14 / 8.28& 0.425 / 0.545& 0.565 / 3.276& 0.239 / 0.381& 8.79 / 14.34 \\
        & OpenCity & 17.90 / 28.80& 24.78 / 40.41& 44.43 / 65.47& 32.16 / 48.47& 2.77 / 6.08& 0.513 / 0.710& 0.412 / 1.740& 0.414 / 0.660& 4.18 / 8.33 \\
      \midrule
      
      \multirow{2}{*}{\shortstack{\textbf{Ours}}}
        & FactoST
          & 17.54 / 28.10& 23.93 / 37.44& 26.48 / 41.92& 18.94 / 29.59
          & 1.96 / 4.51& 0.272 / 0.424& \secondres{0.374} / \secondres{0.545}& \secondres{0.087} / \secondres{0.276}& 4.73 / 12.44 \\
        & \textbf{\method}
          & \boldres{16.75} / \boldres{27.20}& \boldres{22.61} / \secondres{35.95}
          & \boldres{24.70} / 40.79& \boldres{17.65} / \boldres{28.44}
          & \boldres{1.85} / 4.90& \secondres{0.251} / \secondres{0.405}
          & \boldres{0.354} / \boldres{0.540}& \boldres{0.087} / \boldres{0.272}& 4.44 / 12.16 \\
      \midrule
      \multicolumn{11}{c}{\textit{Long-term Forecasting ($96\rightarrow96$)}} \\
      \midrule
      
      \multirow{5}{*}{\shortstack{\textbf{TSEMs}}}
        & Informer & 46.27 / 69.41& 54.26 / 83.42& 52.82 / 81.78& 44.25 / 68.43& 3.27 / 6.81& 2.960 / 3.783& 1.693 / 16.536& 2.249 / 3.403& 6.38 / 12.29 \\
        & FeDFormer & 71.02 / 113.29& 93.52 / 124.07& 74.74 / 108.02& 108.39 / 143.49& 4.20 / 7.34& 1.139 / 1.821& 0.729 / 4.183& 1.336 / 2.091& 9.71 / 16.09 \\
        & DLinear & 76.41 / 113.63& 85.61 / 125.44& 106.68 / 147.43& 76.77 / 109.15& 4.62 / 9.52& 1.069 / 1.781& 0.558 / 3.262& 0.731 / 1.424& 7.65 / 13.42 \\
        & PatchTST & 61.22 / 100.33& 70.71 / 104.00& 80.09 / 118.54& 57.31 / 87.33& 4.32 / 9.22& 0.943 / 1.609& 0.442 / 2.747& 0.708 / 1.409& 7.20 / 14.17 \\
        & TimeMixer & 47.86 / 71.52& 58.44 / 86.67& 67.75 / 100.45& 45.10 / 65.53& 4.11 / 8.94& 1.013 / 1.646& 0.459 / 2.833& 0.720 / 1.401& 7.00 / 13.33 \\
      \midrule
      
      \multirow{5}{*}{\shortstack{\textbf{STEMs}}}
        & GWNet & OOM / OOM& OOM / OOM& OOM / OOM& OOM / OOM& OOM / OOM& OOM / OOM& OOM / OOM& OOM / OOM& OOM / OOM \\
        & D2STGNN & OOM / OOM& OOM / OOM& OOM / OOM& OOM / OOM& OOM / OOM& OOM / OOM& OOM / OOM& OOM / OOM& OOM / OOM \\
        & STID & 45.45 / 65.35& 78.13 / 111.12& 71.32 / 106.95& 75.87 / 103.15& 3.10 / 6.63& 1.066 / 1.751& 0.440 / 2.755& 0.740 / 1.446& \boldres{5.94} / \secondres{11.91} \\
        & STAEformer & 77.42 / 115.67& 64.12 / 91.95& 61.45 / 91.06& 68.45 / 96.14& 3.28 / 6.65& 1.295 / 1.918& 0.733 / 6.562& 1.171 / 1.804& \secondres{6.15} / \boldres{11.56} \\
        & BigST & 51.87 / 75.56& 52.37 / 80.23& 54.92 / 82.12& 58.68 / 86.76& \secondres{2.93} / \secondres{6.20}& 1.164 / 1.797& 0.481 / 2.833& 0.892 / 1.522& 6.69 / 12.22 \\
      \midrule
      
      \multirow{3}{*}{\shortstack{\textbf{TSFMs}}}
        & TimesFM & 38.47 / 59.77& 64.43 / 93.44& 157.10 / 208.36& 89.93 / 125.27& 5.18 / 9.97& 0.365 / 0.541& 0.305 / 0.465& 0.270 / 0.484& 14.23 / 22.56 \\
        & Moirai & 51.40 / 79.47& 81.30 / 116.26& 134.46 / 200.30& 68.73 / 97.89& 5.78 / 10.97& \boldres{0.325} / \boldres{0.465}& 0.312 / 0.484& 0.262 / 0.465& 12.17 / 22.39 \\
        & Rose & 76.67 / 108.90& 100.22 / 137.06& 137.43 / 181.66& 82.34 / 112.47& 4.55 / 9.18& 0.370 / 0.576& 0.273 / 0.433& \secondres{0.224} / 0.434& 8.49 / 21.24 \\
      \midrule
      
      \multirow{3}{*}{\shortstack{\textbf{STFMs}}}
        & UniST
          & 67.70 / 94.00& 85.14 / 112.11& 101.20 / 134.98& 73.81 / 96.45
          & 5.17 / 8.27& 0.488 / 0.622& 0.494 / 2.512& 0.348 / 0.491& 13.16 / 19.96 \\
        & OpenCity
          & 34.21 / 54.82& 67.24 / 112.20& 50.70 / 78.36& 49.47 / 82.07
          & 7.40 / 12.38& 0.751 / 1.040& 0.303 / 1.240& 0.653 / 3.730& 9.71 / 13.62 \\
      \midrule
      
      \multirow{2}{*}{\shortstack{\textbf{Ours}}}
        & FactoST
          & \secondres{28.57} / \secondres{46.78}& \secondres{42.04} / \secondres{64.89}& \secondres{45.60} / \secondres{72.47}& \boldres{38.12} / \boldres{59.72}
          & 2.96 / 6.21& 0.358 / 0.561& \secondres{0.265} / \secondres{0.409}& 0.226 / \secondres{0.426}& 6.93 / 13.07 \\
        & \textbf{\method}
          & \boldres{28.41} / \boldres{46.61}& \boldres{39.73} / \boldres{61.54}
          & \boldres{43.55} / \boldres{72.46}& \secondres{38.48} / \secondres{64.39}
          & \boldres{2.59} / \boldres{5.99}& \secondres{0.328} / \secondres{0.535}
          & \boldres{0.241} / \boldres{0.390}& \boldres{0.207} / \boldres{0.414}& 6.39 / 20.81 \\
      \bottomrule
    \end{tabular}}
  \vspace{-0.5em}
\end{table*}

Expert models are implemented in \texttt{BasicTS}\footnote{\url{https://github.com/GestaltCogTeam/BasicTS}}\cite{liang2022basicts}, TSFMs/STFMs are evaluated with official codebases/checkpoints. Hyperparameters follow original defaults. As prevailing ST models are mostly point predictors, we report MAE/RMSE following~\cite{yuan2024unist}for fair comparison; probabilistic capabilities are demonstrated in \ref{vis_example}.

\begin{figure}[b!]
    \centering
    \includegraphics[width=\linewidth]{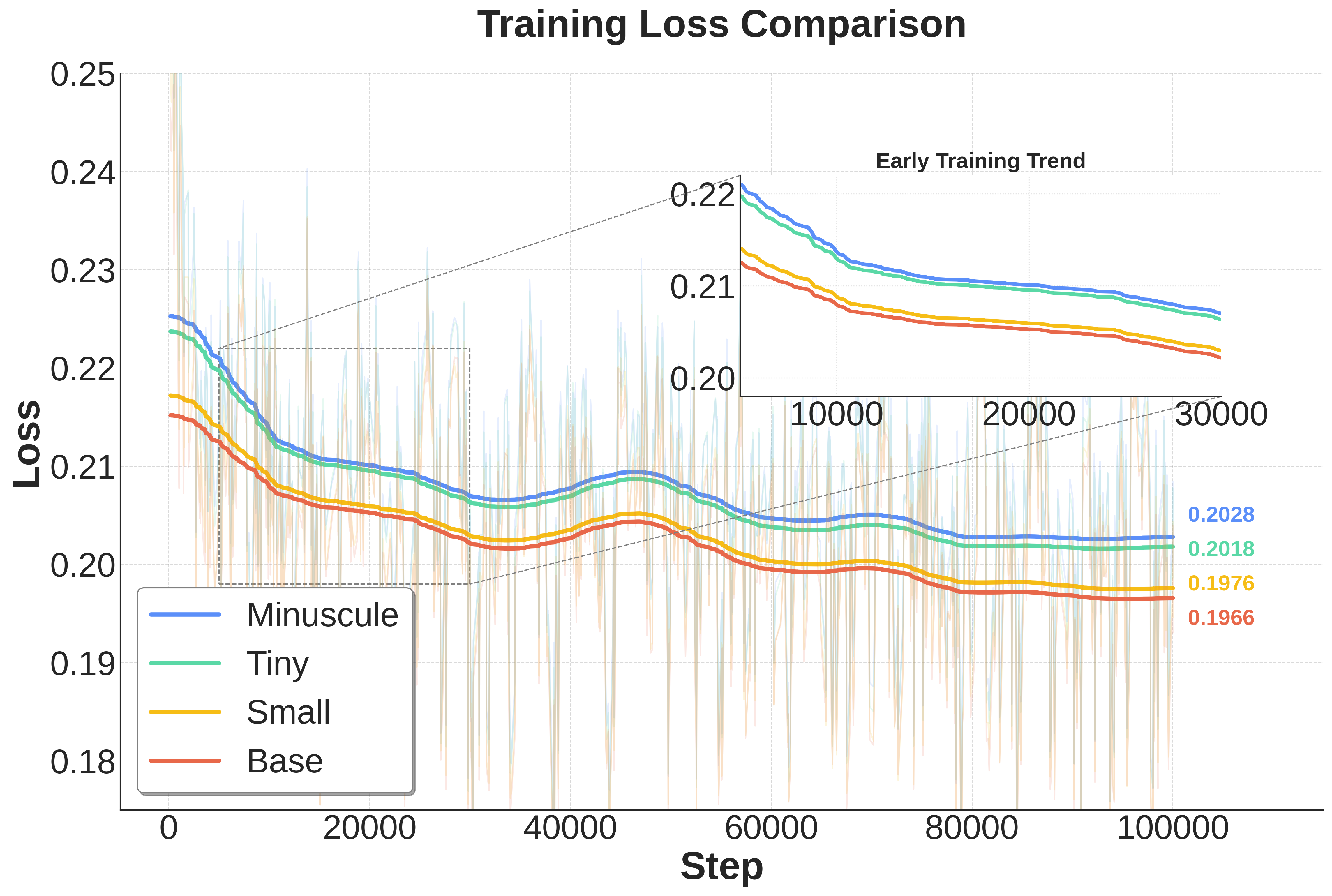}
    \caption{Training loss comparison across different model scales. The zoomed-in window details the early training phase. The final converged loss values are annotated on the right, showing a monotonic improvement in performance as model capacity increases from Minuscule to Base.}
    \label{fig:loss_comparison}
\end{figure}

\vspace{0.3em}
\noindent\textbf{Implementation Details.} 
We implement \method via PyTorch on four NVIDIA A800 (80GB) GPUs. 
Unless otherwise specified, we adopt the \textbf{Tiny} configuration as the default backbone to balance the overall performance and computational efficiency. 
This architecture features a customized Transformer encoder-only structure consisting of 3 layers, 4 attention heads, a latent dimension of $d=256$, and a feed-forward network hidden dimension of 1024. 
To systematically investigate the impact of model capacity, we also design additional variants (Minuscule, Small, Base); detailed hyper-parameter specifications for all variants are provided in Table~\ref{tab:model_variants}.
Sequences are tokenized with a patch size of 16, and dropout rate is set to 0.1. \textbf{For temporal pretraining}, the model is optimized via Adam with a learning rate of $5 \times 10^{-4}$, regulated by a Warmup-Stable-Decay scheduler \cite{wen2024understanding, hu2024minicpm}. The max input sequence length and prediction horizon are set to 2048 and 256 separately, and a batch size of 512 is used to facilitate efficient training on large-scale corpora. \textbf{For ST adaptation}, the learning rate is adjusted to $10^{-3}$, the number of alignment prompt tokens is set to $K = 3$, and the patching configuration remains unchanged from pretraining. For the memory module, the replacement ratio is set to 0.3 and the buffer size to 20\% of the dataset capacity. The ST identifier embedding dimension is set to 32, with a maximum time-lag step of $\Delta = 3$.

\begin{table*}[t!]
  \centering
  \caption{Full-shot Forecasting Comparison. \boldres{Red}=best, \secondres{Blue}=second best. Values: MAE/RMSE. OOM=out-of-memory (NVIDIA A800-80G).}
  \setlength{\tabcolsep}{2pt}
  \footnotesize
  \renewcommand{\arraystretch}{0.8}
  \scalebox{0.95}{
    \begin{tabular}{l|cccccccccc}
      \toprule
      \textbf{Type} & \textbf{Model} 
        & \textbf{PEMS-03} & \textbf{PEMS-04} & \textbf{PEMS-07} & \textbf{PEMS-08} 
        & \textbf{PEMS-Bay} & \textbf{ETTh2} & \textbf{Electricity} & \textbf{Weather} & \textbf{METR-LA} \\
      \midrule
      \multicolumn{11}{c}{\textit{Short-term Forecasting ($12\rightarrow12$)}} \\
      \midrule
      
      \multirow{2}{*}{\shortstack{\textbf{TSEMs}}} 
        & PatchTST & 20.97 / 33.54 & 27.30 / 42.83 & 30.37 / 47.43 & 21.41 / 34.23 & 2.06 / 5.02 & 0.269 / 0.423 & 0.525 / 0.773 & 0.088 / 0.291 & 4.38 / 9.73 \\
        & DLinear & 21.37 / 34.50 & 27.99 / 43.90 & 31.70 / 49.40 & 22.43 / 35.44 & 2.14 / 5.01 & 0.273 / 0.425 & 0.574 / 0.793 & 0.098 / 0.316 & 4.82 / 10.16 \\
      \midrule
      
      \multirow{2}{*}{\shortstack{\textbf{STEMs}}} 
        & D2STGNN & \boldres{14.91} / \secondres{25.82} & \boldres{18.75} / \secondres{30.12} & \boldres{20.19} / \boldres{33.25} & \boldres{14.63} / \boldres{23.73} & \boldres{1.56} / \boldres{3.62} & 0.386 / 0.573 & \boldres{0.259} / \boldres{0.399} & 0.157 / 0.323 & \secondres{3.73} / \boldres{7.21} \\
        & GWNet & 15.93 / 28.11 & 20.93 / 32.96 & 23.86 / 37.83 & 16.48 / 26.19 & 2.00 / 4.55 & \secondres{0.255} / \secondres{0.398} & 0.297 / 0.447 & \boldres{0.084} / \boldres{0.271} & \boldres{3.60} / \secondres{7.41} \\
      \midrule
      
      \multirow{2}{*}{\shortstack{\textbf{Ours}}} 
        & FactoST & 17.54 / 28.10 & 23.93 / 37.44 & 26.48 / 41.92 & 18.94 / 29.59 & 1.96 / 4.51 & 0.272 / 0.424 & 0.374 / 0.545 & 0.087 / 0.276 & 4.74 / 10.08 \\
        & \textbf{\method} & \secondres{15.65} / \boldres{24.90} & \secondres{20.61} / \boldres{32.81} & \secondres{21.95} / \secondres{35.39} & \secondres{15.80} / \secondres{25.46} & \secondres{1.85} / \secondres{4.26} & \boldres{0.238} / \boldres{0.383} & \secondres{0.299} / \secondres{0.443} & \secondres{0.090} / \secondres{0.278} & 4.68 / 9.87 \\
      \midrule
      \multicolumn{11}{c}{\textit{Long-term Forecasting ($96\rightarrow96$)}} \\
      \midrule
      
      \multirow{2}{*}{\shortstack{\textbf{TSEMs}}} 
        & PatchTST & 37.32 / 59.23 & 53.14 / 82.26 & 56.95 / 74.45 & 38.13 / 62.86 & 3.95 / 8.49 & \boldres{0.334} / \boldres{0.538} & \secondres{0.261} / \secondres{0.437} & \secondres{0.212} / \secondres{0.422} & \boldres{7.19} / 14.37 \\
        & DLinear & 69.17 / 103.71 & 82.72 / 120.04 & 105.40 / 144.38 & 75.53 / 106.20 & 4.47 / 9.17 & 0.345 / 0.546 & 0.273 / 0.453 & 0.237 / 0.459 & 7.85 / 13.50 \\
      \midrule
      
      \multirow{2}{*}{\shortstack{\textbf{STEMs}}} 
        & D2STGNN & OOM / OOM & OOM / OOM & OOM / OOM & OOM / OOM & OOM / OOM & OOM / OOM & OOM / OOM & OOM / OOM & OOM / OOM \\
        & GWNet & OOM / OOM & OOM / OOM & OOM / OOM & OOM / OOM & OOM / OOM & OOM / OOM & OOM / OOM & OOM / OOM & OOM / OOM \\
      \midrule
      
      \multirow{2}{*}{\shortstack{\textbf{Ours}}} 
        & FactoST & 28.57 / 46.78 & 42.04 / 64.89 & 45.60 / 72.47 & 35.69 / 56.15 & 2.96 / 6.21 & 0.364 / 0.564 & 0.265 / 0.409 & 0.226 / 0.426 & 8.17 / \secondres{13.69} \\
        & \textbf{\method} & \boldres{25.05} / \boldres{42.77} & \boldres{26.68} / \boldres{42.36} & \boldres{28.30} / \boldres{46.35} & \boldres{19.96} / \boldres{32.36} & \boldres{2.75} / \boldres{5.40} & \secondres{0.358} / \secondres{0.561} & \boldres{0.227} / \boldres{0.367} & \boldres{0.209} / \boldres{0.402} & \secondres{7.40} / \boldres{13.28} \\
      \midrule
    \end{tabular}}
  \label{tab:full_shot_forecasting}
  \vspace{-0.5em}
\end{table*}
\begin{table*}[t!]
  \centering
  \caption{Zero-shot forecasting of foundation models. *: Pretraining-seen datasets (excluded from ranking). Values: MAE/RMSE.}
  \setlength{\tabcolsep}{4.5pt}
  \footnotesize
  \renewcommand{\arraystretch}{0.8}
  \resizebox{0.98\linewidth}{!}{
    \begin{tabular}{lc|cc|cc|ccc}
      \toprule
      \multicolumn{2}{c|}{\textbf{Model Type}} 
        & \multicolumn{2}{c|}{\textbf{Ours}}
        & \multicolumn{2}{c|}{\textbf{STFMs}}
        & \multicolumn{3}{c}{\textbf{TSFMs}} \\
      \cmidrule{1-2}\cmidrule{3-4}\cmidrule{5-6}\cmidrule{7-9}
      \textbf{Dataset} & \textbf{Horizon}
        & \textbf{FactoST-V2}
        & \textbf{FactoST}
        & \textbf{OpenCity}
        & \textbf{UniST}
        & \textbf{Rose}
        & \textbf{TimesFM}
        & \textbf{Moirai} \\
      \midrule
      \multirow{2}{*}{PEMS-03}
       & 12→12 & \boldres{25.70} / \boldres{40.75} & 30.12 / \secondres{46.92} & 30.37 / 47.49 & 101.87 / 129.94 & 29.74 / 47.27 & \secondres{29.71} / 49.21 & 28.23* / 46.36* \\
       & 96→96 & \boldres{86.90} / \boldres{124.62} & 113.62 / 144.80 & 125.18 / 159.23 & \secondres{102.87} / \secondres{137.90} & 122.93 / 158.76 & 108.59 / 152.74 & 74.85* / 110.78* \\
      \midrule
      \multirow{2}{*}{PEMS-04}
       & 12→12 & \boldres{33.13} / \boldres{49.92} & 38.65 / 56.59 & 39.34* / 58.50* & 67.91 / 87.63 & 38.35 / 57.34 & \secondres{35.00} / \secondres{53.27} & 34.65* / 52.13* \\
       & 96→96 & \boldres{107.15} / \boldres{147.60} & 142.11 / 175.99 & 153.18* / 188.95* & \secondres{115.76} / \secondres{153.58} & 151.39 / 189.29 & 127.39 / 171.13 & 105.95* / 141.40* \\
      \midrule
      \multirow{2}{*}{PEMS-Bay}
       & 12→12 & \secondres{2.27} / \secondres{5.46} & \boldres{2.02} / \boldres{4.59} & 3.23* / 6.91* & 14.89 / 16.92 & 2.72 / 6.41 & 6.59 / 14.99 & 1.97* / 4.69* \\
       & 96→96 & \boldres{6.11} / \secondres{11.56} & \secondres{6.12} / \boldres{10.47} & 6.92* / 11.79* & 9.29 / 12.72 & 7.02 / 11.90 & 14.16 / 24.01 & 6.51* / 11.70* \\
      \midrule
      \multirow{2}{*}{PEMS-07}
       & 12→12 & \boldres{37.85} / \boldres{57.48} & 45.47 / 66.68 & 45.18 / 67.15 & 104.93 / 133.65 & 44.79 / 67.33 & \secondres{42.10} / \secondres{64.59} & 35.64* / 50.25* \\
       & 96→96 & \boldres{126.83} / \boldres{173.31} & 156.07 / 190.59 & 172.20 / 211.25 & \secondres{144.67} / \secondres{184.62} & 172.07 / 212.94 & 151.99 / 205.18 & 125.91* / 169.43* \\
      \midrule
      \multirow{2}{*}{PEMS-08}
       & 12→12 & \boldres{27.11} / \boldres{41.18} & 32.14 / 47.27 & 32.45* / 48.42* & 73.46 / 93.14 & 31.70 / 47.77 & \secondres{29.68} / \secondres{45.18} & 38.23* / 53.12* \\
       & 96→96 & \boldres{91.09} / \secondres{127.28} & 121.43 / 151.59 & 128.48* / 161.75* & 104.77 / 136.21 & 129.05 / 163.15 & \secondres{92.72} / \boldres{126.10} & 119.10* / 151.83* \\
      \midrule
      \multirow{2}{*}{METR-LA}
       & 12→12 & \boldres{5.32} / \boldres{12.82} & 5.98 / 14.26 & 4.30* / 8.37* & 24.33 / 29.31 & 6.05 / 14.36 & 6.59 / 14.99 & \secondres{5.55} / \secondres{13.79} \\
       & 96→96 & \boldres{10.87} / \secondres{21.49} & \secondres{12.68} / \boldres{18.82} & 10.85* / 17.60* & 25.88 / 30.19 & 12.96 / 23.61 & 14.16 / 24.01 & 12.87 / 22.19 \\
      \bottomrule
    \end{tabular}
  }
  \label{tab:zero_shot_comparison}
  \vspace{-1em}
\end{table*}

\subsection{Few-shot Prediction}

\vspace{0.3em}
\noindent\textbf{Setting.} We evaluate \method under a realistic few-shot adaptation scenario, where only the final 10\% of the labeled training data is used for fine-tuning. Following established protocols~\cite{shao2022decoupled, shao2022spatial}, we assess performance on both short-term ($12 \to 12$) and long-term ($96 \to 96$) forecasting horizons to examine the model's capacity for rapid adaptation to new domains with minimal supervision.

\noindent\textbf{Results.} As presented in \autoref{tab:few_shot_forecasting}, \method establishes state-of-the-art performance across both horizons, empirically validating the superiority of the factorized design. Unlike coupled STFMs (e.g., UniST~\cite{yuan2024unist}, OpenCity~\cite{li2024opencity}) that risk negative transfer by encoding rigid source-domain spatial priors, our decoupled paradigm mitigates such inductive conflicts, allowing the lightweight STA module to rapidly align with target-domain topologies using limited supervision. Simultaneously, \method transcends pure TSFMs (e.g., TimesFM~\cite{das2024decoder}) by explicitly modulating localized ST contexts, demonstrating that scaling temporal capacity alone is insufficient for modeling complex inter-node dependencies. Notably, this advantage is amplified in long-term forecasting; while traditional STEMs (e.g., D2STGNN~\cite{shao2022decoupled}) suffer from memory bottlenecks due to quadratic $\mathcal{O}(N^2)$ complexity, \method maintains robust accuracy with linear efficiency, proving its scalability for resource-constrained deployment.

\subsection{Full-shot Prediction}
\vspace{0.3em}
\noindent\textbf{Setting.} We evaluate \method under the full-shot setting using the complete training set. This scenario rigorously assesses the model’s \textit{asymptotic capacity} and computational scalability when data scarcity is no longer the primary constraint.

\noindent\textbf{Results.} As presented in Table~\ref{tab:full_shot_forecasting}, \method achieves a superior trade-off between accuracy and scalability compared to task-specific baselines. The experiments first expose a critical scalability barrier in coupled ST architectures (e.g., GWNet, D2STGNN), which succumb to memory exhaustion (OOM) in long-horizon forecasting due to quadratic complexity and susceptibility to \textit{spatio-temporal over-squashing}. \method surmounts these hurdles via its factorized linear-complexity design, validating that spatial decoupling is prerequisite for scaling foundation models to large graphs and long time steps. Furthermore, the results underscore the dichotomy between inductive generalization and transductive overfitting. While specialized expert models degrade significantly in cross-domain settings (e.g., Energy, Weather) due to overfitting static graph topologies, \method acquires inductive spatial meta-patterns through adaptation. This confers substantial robustness, consistently ranking within the top-2 across diverse non-traffic domains where pure time-series models (e.g., PatchTST) fail due to the absence of spatial reasoning.

\subsection{Zero-shot Prediction}

\vspace{0.3em}
\noindent\textbf{Setting.} We evaluate zero-shot forecasting over short and long horizons to quantify foundation models' generalization of representations to unseen domains without parameter updates.

\vspace{0.3em}
\noindent\textbf{Results.} As evidenced in Table~\ref{tab:zero_shot_comparison}, \method demonstrates robust zero-shot generalization, effectively rivaling or surpassing large-scale foundation models. The results illuminate a critical insight: coupled STFMs (e.g., OpenCity) frequently underperform spatially-agnostic baselines (e.g., TimesFM), suffering from the \textit{curse of coupled spatial priors} where rigid source-domain topologies induce negative transfer to dissimilar target distributions. \method circumvents this by strictly decoupling spatial adaptation, empirically validating the hypothesis that while temporal dynamics (e.g., periodicity, trends) adhere to universal physical laws, spatial correlations are predominantly domain-specific artifacts requiring isolation. Consequently, our minimalist UTP design—empowered by randomized sequence masking and quantile objectives—achieves superior cross-domain transfer by prioritizing the distillation of invariant temporal representations over joint spatio-temporal overfitting, proving that architectural disentanglement is a more decisive factor than brute-force modeling for zero-shot scenarios.

\subsection{Ablation Studies}

\vspace{0.3em}
\noindent\textbf{UTP Architectural Ablation.}
To verify the architectural choices in UTP stage, we conducted a detailed ablation study on the PEMS08 dataset under zero-shot setting (see \autoref{tab:ablation}). First, the inclusion of \textit{p-RoPE} yields consistent gains (+0.62\% MAE) over vanilla RoPE. By selectively applying rotary embeddings to high-frequency components while preserving low-frequency trend semantics, p-RoPE ensures that sequential position modeling does not compromise signal amplitude fidelity. Furthermore, the degradation observed upon replacing \textit{Gated Attention} with standard mechanisms (+3.97\% MAE) validates the necessity of dynamic noise filtering to mitigate "attention sinks" inherent in raw time series. The exclusion of \textit{Quantile Loss} also leads to a notable decline (+7.01\% MAE), underscoring its role in enhancing forecast robustness by explicitly modeling the full conditional distribution rather than just point estimates. Finally, the results highlight \textit{Random Sequence Masking} as the cornerstone of universality; its removal causes a catastrophic performance drop (+17.71\% MAE), confirming that exposing the backbone to variable effective contexts is a prerequisite for learning length-agnostic temporal laws suitable for diverse downstream horizons.

\begin{table}[h]
    \centering
    \caption{Ablation study on Gated Attention and p-RoPE on Zero-shot Forecasting (PEMS08, \textit{Long-term Forecasting} 96 $\to$ 96 setting).}
    \label{tab:ablation}
    \scalebox{0.90}{
    \begin{tabular}{lcccc}
        \toprule
        \textbf{Model Variants} & \multicolumn{2}{c}{\textbf{MAE}} & \multicolumn{2}{c}{\textbf{RMSE}} \\
        \cmidrule(lr){2-3} \cmidrule(lr){4-5}
        & \textbf{Value} & \textbf{\%$\Delta$} & \textbf{Value} & \textbf{\%$\Delta$} \\
        \midrule
        \method & \boldres{91.09} &  & \boldres{127.28} &  \\
        w/o p-RoPE (vanilla RoPE) & \secondres{91.65} & +0.62\% $\uparrow$ & \secondres{127.76} & +0.38\% $\uparrow$ \\
        w/o Gated Attn (vanilla Attn) & 94.71 & +3.97\% $\uparrow$ & 131.80 & +3.55\% $\uparrow$ \\
        w/o Quantile Loss & 97.48 & +7.01\% $\uparrow$ & 135.15 & +6.18\% $\uparrow$ \\
        w/o Random Sequence Mask & 107.22 & +17.71\% $\uparrow$ & 145.72 & +14.49\% $\uparrow$ \\
        \bottomrule
    \end{tabular}}
    \vspace{-1em}
\end{table}

\begin{figure}[h!]
    \centering
    \includegraphics[width=\columnwidth]{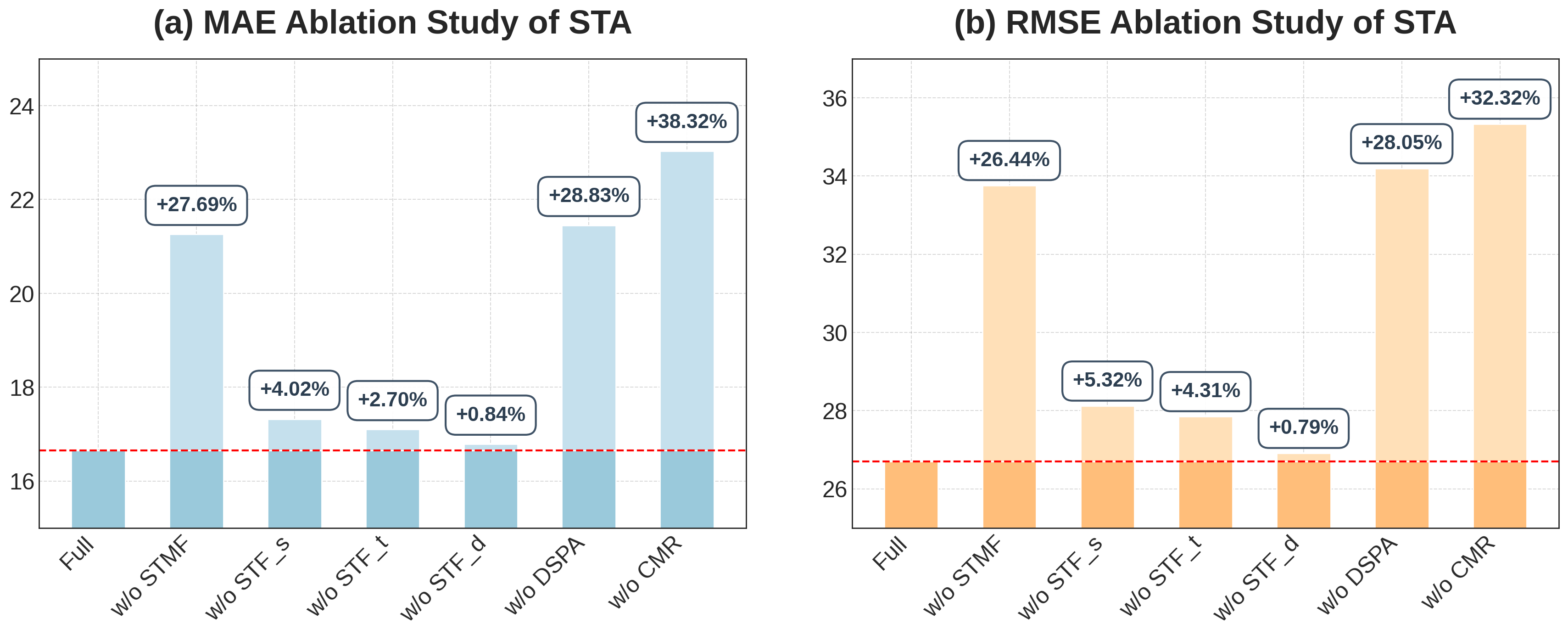}
    \caption{Ablation of STA's components on PEMS-03 short-term forecasting, showing MAE/RMSE degradation upon removing key modules.}
    \label{fig:ablation}
    \vspace{-1em}
\end{figure}

\begin{figure}[b!]
    \centering
    \includegraphics[width=\columnwidth]{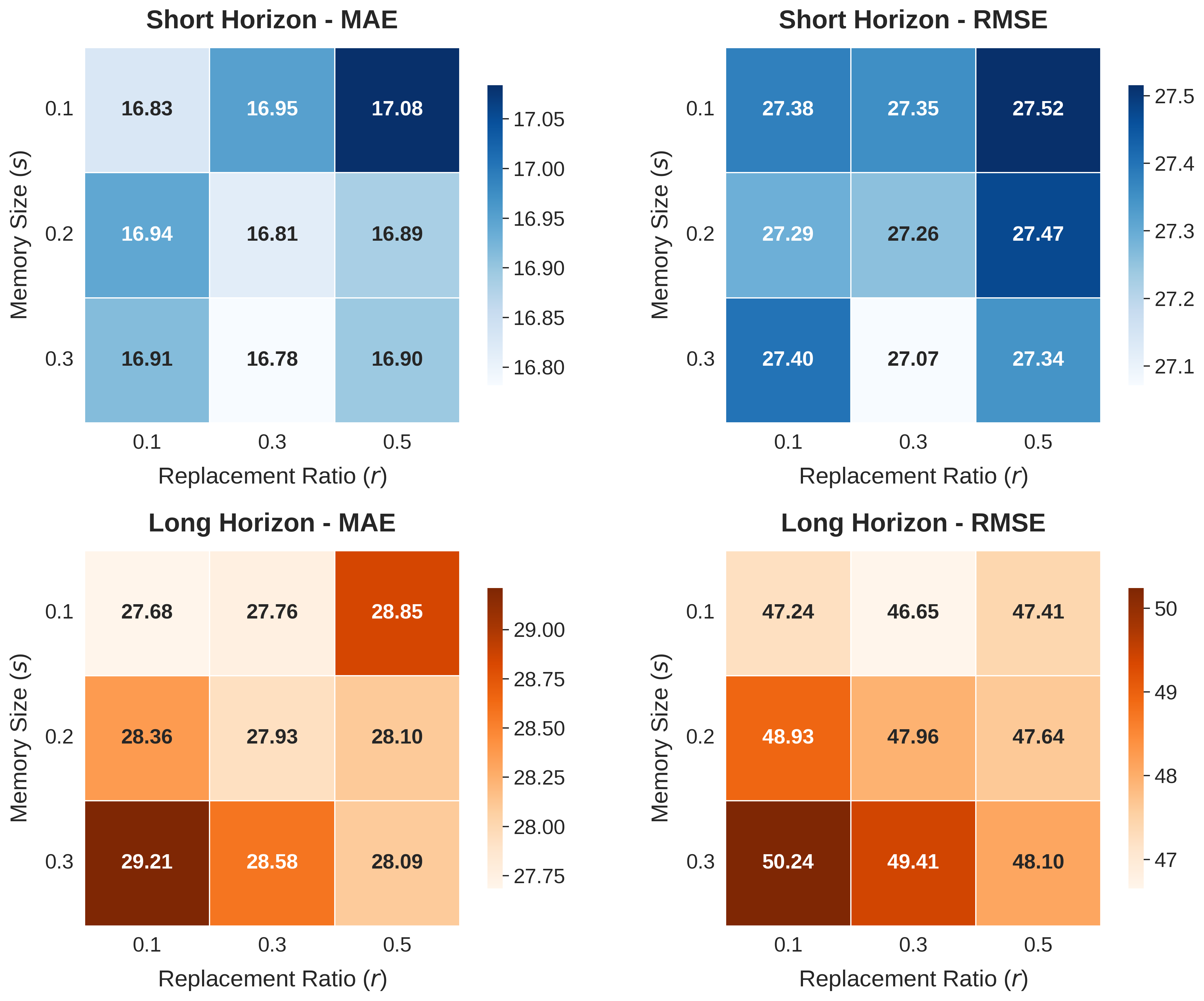} 
    \caption{Parameter sensitivity analysis of CMR on PEMS-03. The heatmaps of MAE/RMSE under varying memory sizes ($s$) and replacement ratios ($r$) exhibit a flat landscape, demonstrating the parameter robustness of CMR.}
    \label{fig:cmr_sensitivity}
\end{figure}

\vspace{0.3em}
\noindent\textbf{STA's Component Ablation}. 
We evaluate the contribution of each STA module on the PEMS-03 dataset. As detailed in Figure~\ref{fig:ablation}, removing any component leads to performance degradation, revealing a clear hierarchical dependency. \textbf{First}, \textit{STMF} is fundamental to ST-aware modeling: its removal leads to a substantial error increase around 28\%, validating its role in providing essential multi-granular context. This foundation is further refined by \textit{STF}; ablation of its affinity components shows spatial affinity ($S_s$) contributes most, followed by temporal ($S_t$) and time-lagged ($S_d$) components. These gaps confirm STF acts as a meta inductive bias mechanism that dynamically reweights ST interactions for dynamic environments. \textbf{Second}, regarding domain alignment, omitting \textit{DSPA} induces a moderate degradation. This confirms that learnable global prompts provide valuable complementary alignment, fine-tuning the frozen backbone to the target manifold with negligible parameter overhead. \textbf{Finally}, \textit{CMR} ensures training stability: its removal causes the most severe performance deterioration (over 30\%), highlighting the model's vulnerability to catastrophic forgetting without distributional anchors. Sensitivity analysis of CMR hyperparameters (memory size $s$, replacement ratio $r$; Figure~\ref{fig:cmr_sensitivity}) reveals a stable performance landscape with minimal fluctuations, confirming CMR's robustness to configuration variations and low tuning overhead.

\subsection{Scaling Analysis}

\begin{figure}[h!]
    \centering
    \includegraphics[width=\columnwidth]{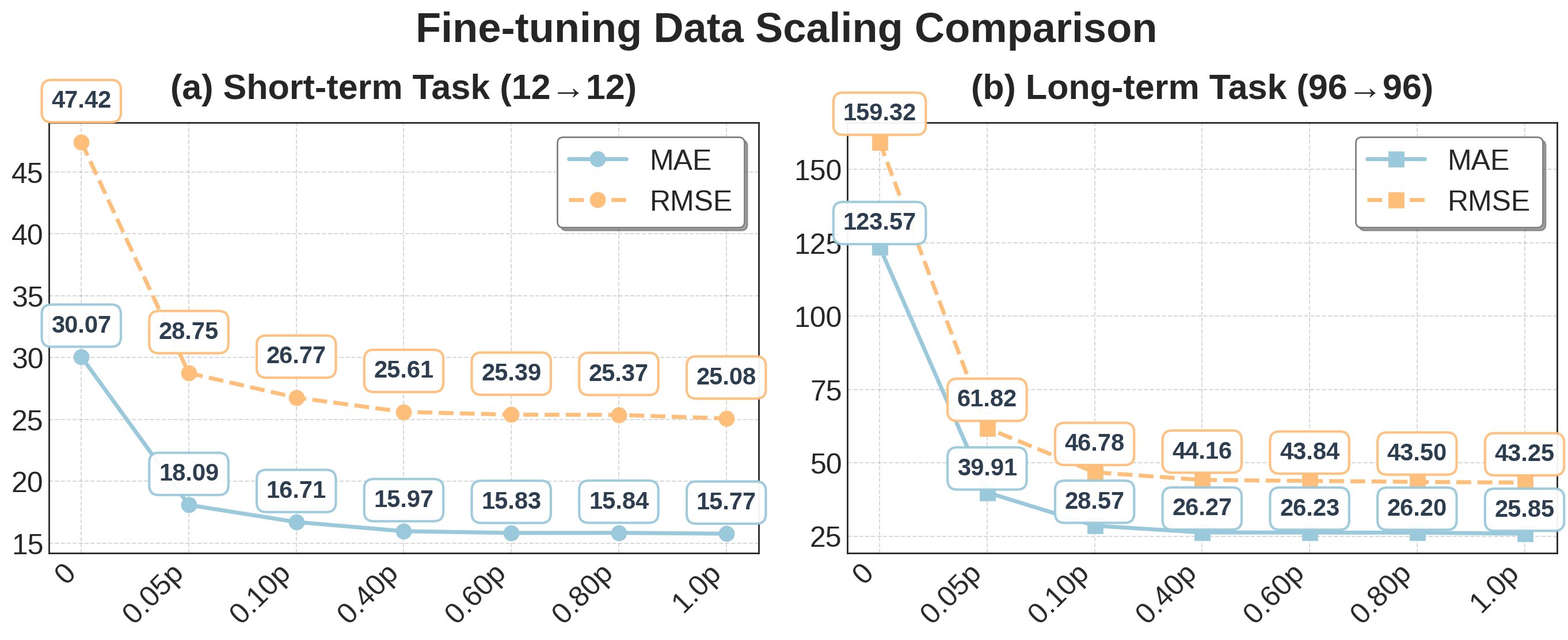}
    \caption{Fine-tuning Data Efficiency Analysis on PEMS-03. The steep learning curves in both short-term (a) and long-term (b) tasks demonstrate a fast adaptation phenomenon: \method matches full-shot training performance with only 10\% labeled data, verifying that its factorized adaptation pipeline enables rapid target-domain alignment even in data-scarce regimes.}
    \label{fig:scaling}
\end{figure}

\vspace{0.3em}
\noindent\textbf{Fine-tuning Data Scaling.} We evaluate the data efficiency of \method on PEMS-03 by varying the fine-tuning ratio from zero-shot (0\%) to full-shot (100\%) (see \autoref{fig:scaling}). The experimental results reveal a striking and consistent \textit{fast adaptation phenomenon}: with a mere 10\% fraction of labeled data, our model achieves highly competitive near-optimal performance, cutting the short-term MAE sharply from approximately 30.1 (zero-shot setting) to 16.7—this value closely approaches the full-shot baseline of 15.77, with a marginal gap of only 0.93. Similarly, long-term error decreases drastically from ~123.6 to ~28.5 at the same 10\% threshold. This remarkably steep convergence behavior strongly validates the efficacy of our proposed factorized model design: the well-pre-trained frozen UTP backbone encodes a robust high-density temporal feature manifold, which enables the compact lightweight adapter to seamlessly and efficiently align with target-domain ST dynamics under minimal supervision, thereby ensuring practical deployment in data-scarce regimes.

\begin{figure}[h!]
    \centering
    \includegraphics[width=\columnwidth]{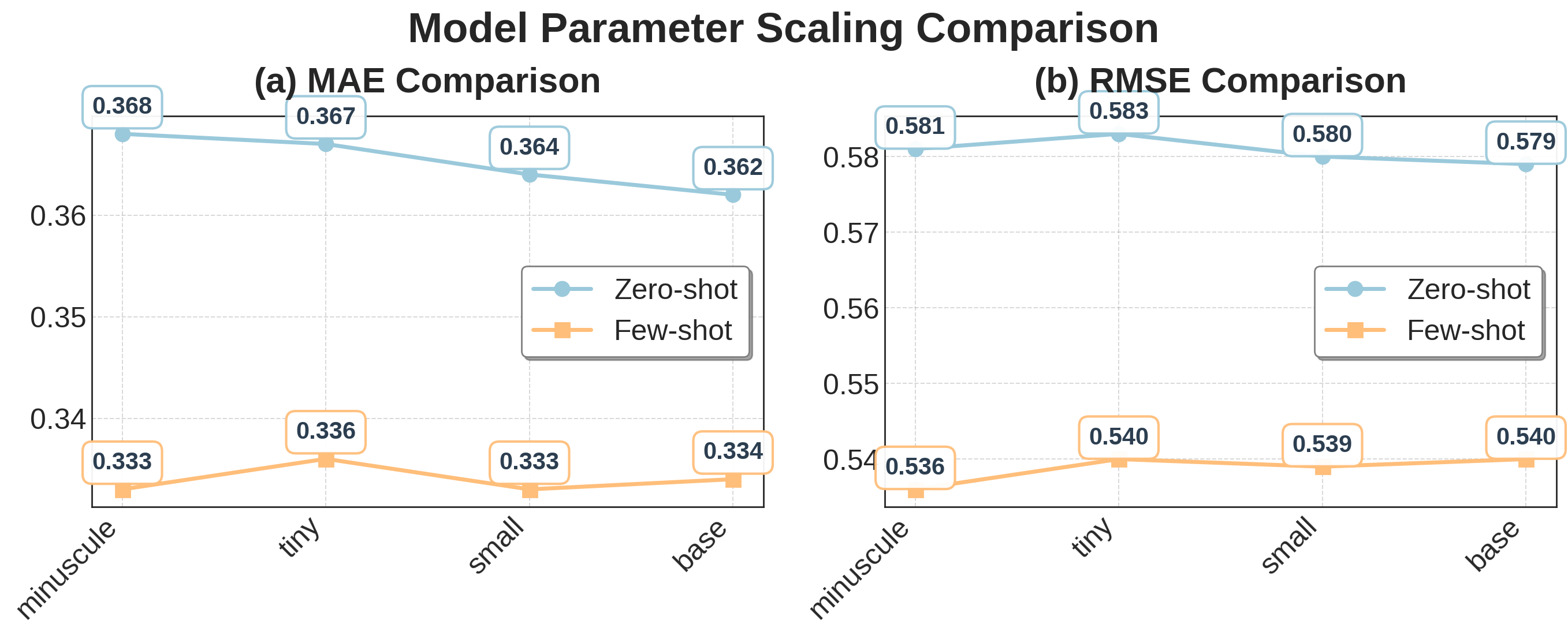}
    \caption{Capacity-Adaptability Trade-off Analysis w.r.t. Backbone Depth on ETTh2. A clear divergence emerges: zero-shot performance (blue) improves with deeper architectures, while few-shot adaptation (orange) does not improve a lot via parameter scaling.}
    \label{fig:model_layers}
\end{figure}

\vspace{0.3em}
\noindent\textbf{Model Size Scaling.}
As shown in Figure~\ref{fig:model_layers}, we explore the \textit{Capacity-Adaptability Trade-off} by varying the model capacity across four variants (Minuscule, Tiny, Small, Base) on ETTh2. A distinct and clear divergence is observed: Zero-shot performance basically and steadily improves with model size (MAE $0.368 \to 0.362$), indicating that larger architectures possess stronger inherent capacity to memorize richer universal patterns. Conversely, few-shot performance remains relatively stable across different scales (MAE fluctuates between $0.333$ and $0.336$) rather than degrading. This stability suggests that for few-shot adaptation, a \textit{moderate model size is sufficient}, as the lightweight adapter effectively aligns the backbone with the target domain even at smaller scales, and further parameter scaling yields diminishing returns for adapted performance.

\subsection{Model Analysis}

\begin{figure}[h!]
    \centering
    \includegraphics[width=1\linewidth]{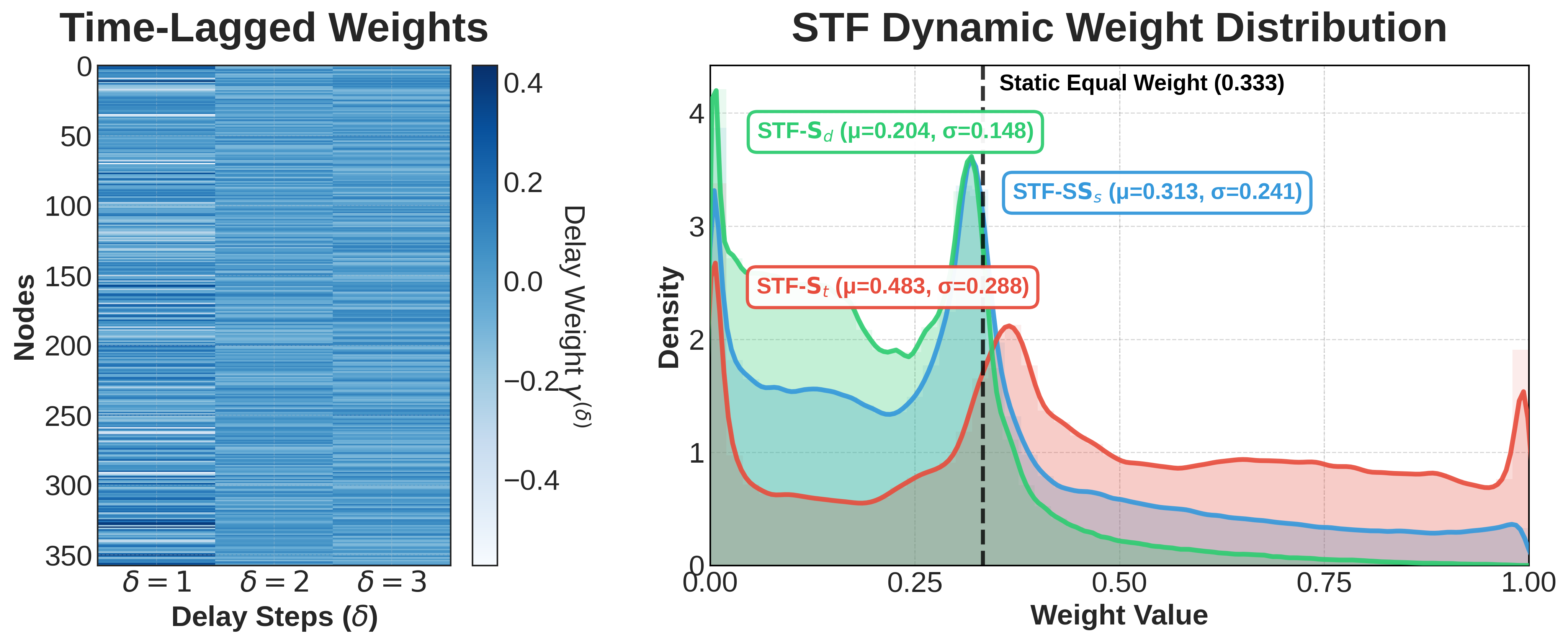}
    \caption{Dynamic ST weighting and learnable delay coefficients in STF. Right: KDEs show context-adaptive weight distributions for $\mathbf{S}_s$, $\mathbf{S}_t$, $\mathbf{S}_d$ vs. static equal weighting. Left: Heatmap of $\gamma^{(\delta)}$ reveals node-specific delay preferences, confirming asynchronous correlation modeling.}
    \label{fig:stf_weight_dist}
\end{figure}

\vspace{0.3em}
\noindent\textbf{STF Interpretability}
To validate the adaptive modeling capability of the STF module, we visualize its core mechanisms on the PEMS-03 short-term forecasting task in Figure~\ref{fig:stf_weight_dist}. The right panel shows kernel density estimates (KDEs) of STF’s dynamic weights for spatial ($\mathbf{S}_s$), temporal ($\mathbf{S}_t$), and time-lagged ($\mathbf{S}_d$) affinities, compared against a static equal-weight baseline (dashed line, value = 0.333). STF exhibits context-adaptive distributions: $\mathbf{S}_t$ has the highest mean ($\mu=0.483$, $\sigma=0.288$), followed by $\mathbf{S}_s$ ($\mu=0.313$, $\sigma=0.241$) and $\mathbf{S}_d$ ($\mu=0.204$, $\sigma=0.148$), confirming that STF automatically adjusts dimension importance without relying on fixed assumptions. The left panel visualizes STF’s learnable delay coefficients $\gamma^{(\delta)}$ as a heatmap across nodes and delay steps ($\delta=1,2,3$). It reveals heterogeneous lag preferences: some nodes prioritize $\delta=2$ (mean $\approx0.54$), while others favor $\delta=3$ (mean $\approx0.6$). Contrary to the original claim, the global mean shows $\delta=1$ as the dominant lag, aligning with short-term traffic dynamics. This verifies that STF captures asynchronous spatio-temporal correlations rather than imposing uniform delays across all locations.

\begin{figure}[h!]
    \centering
    \includegraphics[width=\linewidth]{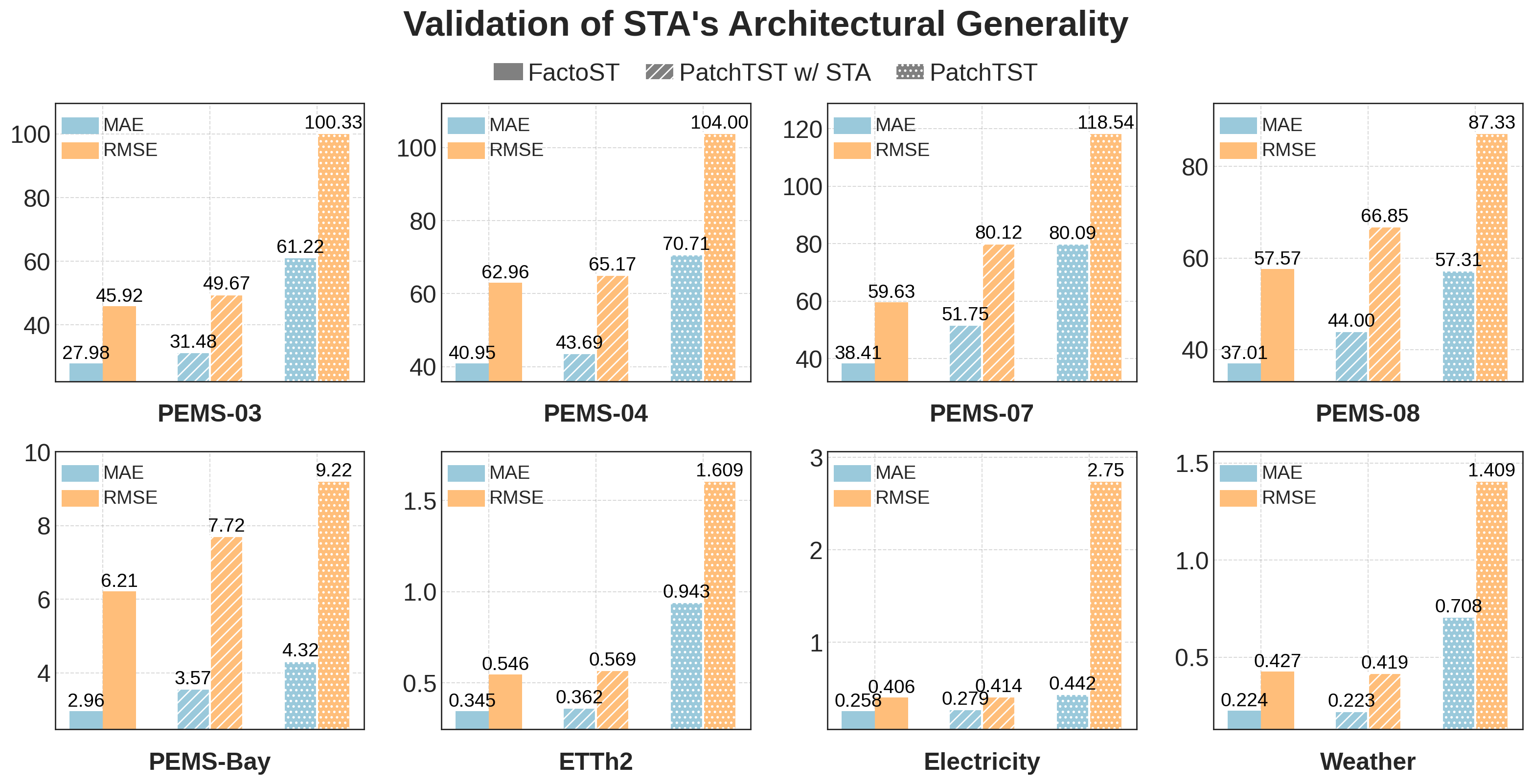}
    \caption{Validation of STA's Architectural Generality for Few-Shot Long-Horizon Forecasting on PEMS03. Integrating STA into PatchTST validates its plug-and-play ability. \textbf{Left:} Consistent gains of "PatchTST w/ STA" over vanilla PatchTST confirm STA injects spatial inductive biases into non-GNN backbones without conflicts. \textbf{Right:} \method still outperforms the enhanced PatchTST, emphasizing that while spatial adaptation is critical, the temporal modeling quality learned by UTP determines the performance upper bound.}
    \label{fig:sta_generality}
\end{figure}

\vspace{0.3em}
\noindent\textbf{Architecture Generality of STA.} 
A pivotal design attribute of the STA module is its \textit{architectural agnosticism}. Unlike traditional STGNN components that are tightly coupled with specific graph topologies, STA operates exclusively on high-level feature embeddings, functioning as a plug-and-play spatial injector. To empirically validate this universality, we integrate STA into PatchTST—a representative channel-independent Transformer that intrinsically lacks spatial modeling capabilities. As illustrated in \autoref{fig:sta_generality}, the \textit{PatchTST+STA} variant yields consistent performance gains over the vanilla PatchTST across few-shot benchmarks. This confirms STA's efficacy in explicitly injecting spatial inductive biases into purely temporal backbones without requiring architectural restructuring. Furthermore, FactoST+ retains a significant performance lead over \textit{PatchTST+STA}. This comparison unveils a critical insight: while STA effectively handles domain-specific spatial adaptation, the model's performance upper bound is fundamentally determined by the quality of the temporal manifold. FactoST+ benefits from UTP on large-scale corpora, demonstrating that \textit{robust temporal representations are the bedrock of cross-domain generalization}, upon which lightweight spatial adaptation can be most effective.

\vspace{0.3em}
\noindent\textbf{Temporal Feature Granularity Analysis.}  
\method enables flexible multi-scale periodicity modeling via its STMF module. Main experiments adopt diverse fine-grained temporal identifiers---\texttt{minute\_of\_hour} (60), \texttt{time\_of\_day} (24), \texttt{day\_of\_week} (7)---to capture high-frequency traffic patterns (e.g., PEMS-03); coarser granularities (monthly/yearly) are readily supported via input feature swapping, no architectural changes required. For granularity evaluation, we replace the original fine-grained features with \texttt{week\_of\_month} (4) and \texttt{month\_of\_year} (12) in a PEMS-03 few-shot setting. As shown in Table~\ref{tab:temporal_granularity}, this induces significant degradation: short-term MAE ↑18.76\%, long-term MAE ↑17.62\%. This confirms \textit{temporal feature design must align with data's intrinsic periodicities}---a core principle inherently supported by our plug-and-play metadata injection mechanism.

\begin{table}[h!]
  \centering
  \caption{Impact of Temporal Feature Granularity on Forecasting Accuracy (PEMS-03). Fine-grained cyclic features (e.g., minute, hour) are indispensable for high-frequency traffic dynamics; coarse granularity causes severe information loss. (\texttt{moh}: minute-of-hour, \texttt{tod}: time-of-day, \texttt{dow}: day-of-week.}
  \resizebox{0.95\linewidth}{!}{
  \begin{tabular}{l c c c c}
    \toprule
    \multirow{2}{*}{\textbf{Feature Set Configuration}} & \multicolumn{2}{c}{\textbf{Short-term Task}} & \multicolumn{2}{c}{\textbf{Long-term Task}} \\
    \cmidrule(lr){2-3} \cmidrule(lr){4-5}
     & \textbf{MAE} & \textbf{RMSE} & \textbf{MAE} & \textbf{RMSE} \\
    \midrule
    Fine-grained \small{(\texttt{moh} + \texttt{tod} + \texttt{dow})} & \boldres{16.68} & \boldres{26.74} & \boldres{27.98} & \boldres{45.92} \\
    Coarse-grained \small{(\texttt{wom} + \texttt{moy})} & \secondres{19.81} & \secondres{31.37} & \secondres{32.91} & \secondres{52.01} \\
    \midrule
    \textit{Error Increase ($\Delta$)} & +18.76\% $\uparrow$ & +17.31\% $\uparrow$ & +17.62\% $\uparrow$ & +13.26\% $\uparrow$ \\
    \bottomrule
  \end{tabular}}
  \label{tab:temporal_granularity}
\end{table}

\begin{figure}[h!]
    \centering
    \includegraphics[width=\columnwidth]{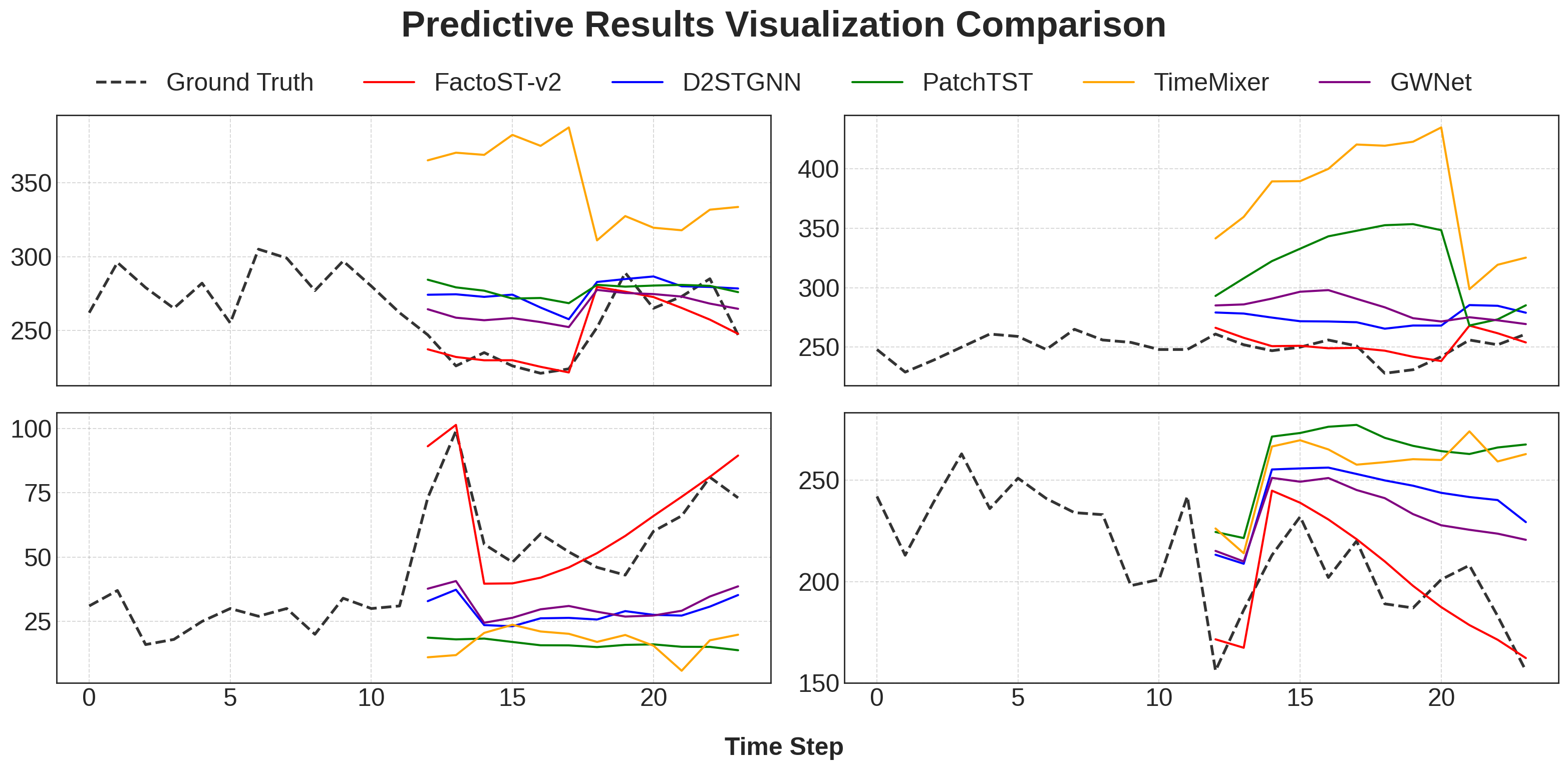}
    \caption{Short-term forecasting curve visualization for four PEMS-03 samples. \method’s prediction closely aligns with the ground truth, effectively capturing both short-term perturbations and periodic trends of traffic flow.}
    \label{fig:pems1212_vis_comparsion}
\end{figure}

\begin{figure}[h!]
    \centering
    \includegraphics[width=\linewidth]{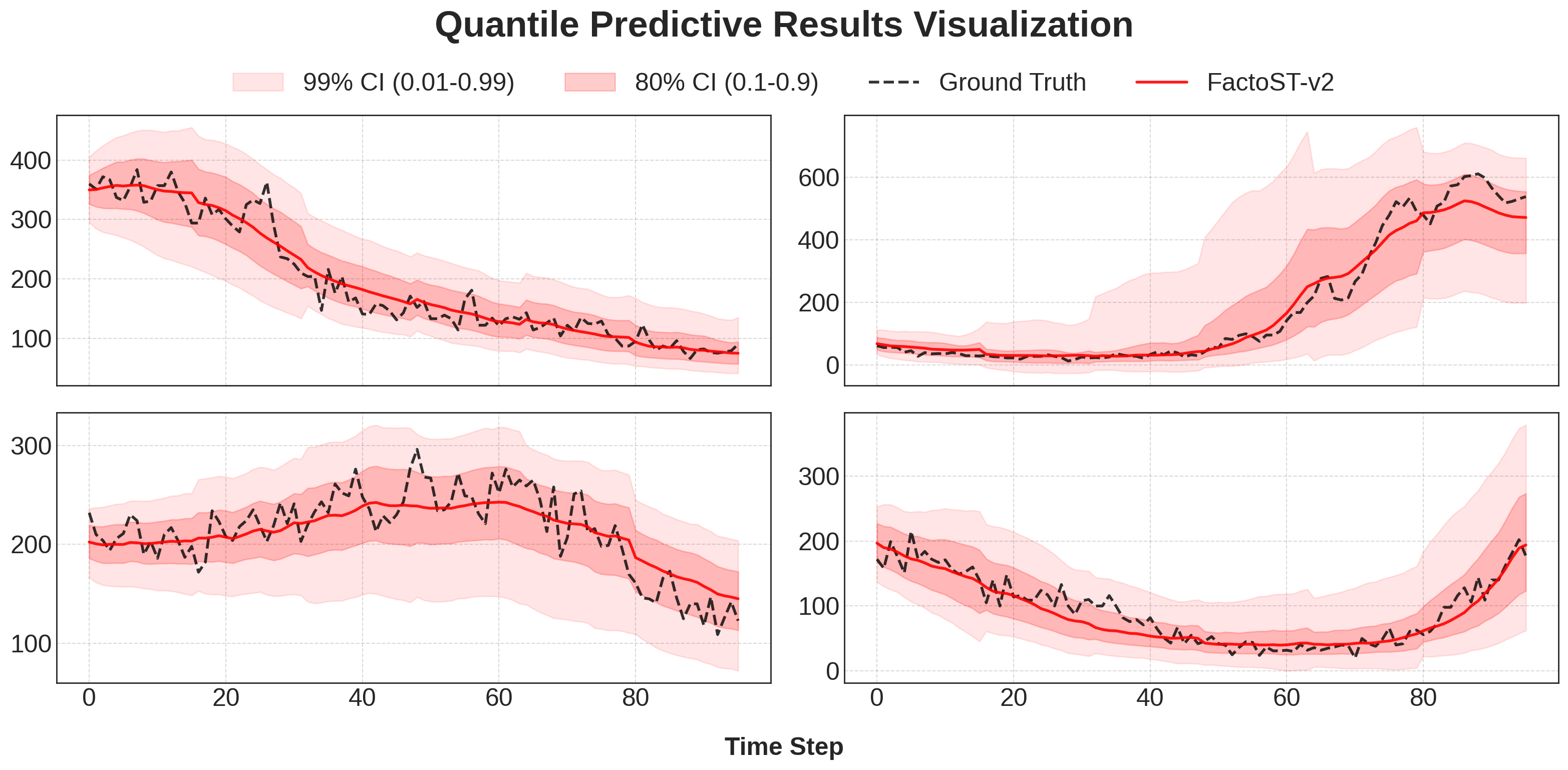}
    \caption{Quantile Forecasting Visualization (PEMS-03). Adaptive confidence intervals (80\%, 99\%) cover ground truth across flow conditions, verifying learned conditional distribution robustness.}
    \label{fig:quantile_prediction_vis}
\end{figure}

\vspace{0.3em}
\noindent\textbf{Visualization Examples.} \label{vis_example}
As shown in \autoref{fig:pems1212_vis_comparsion}, we qualitatively visualize the short-term forecasting curves for four PEMS-03 samples. The ground truth exhibits both distinct abrupt traffic perturbations and inherent periodic fluctuations. \method's prediction aligns most strikingly closely with these dynamics, tightly following sharp short-term changes while precisely matching the exact timing and amplitude of periodic peaks and valleys. In contrast, other models (e.g., D2STGNN, GWNet) show clear noticeable deviations in either tracking perturbations or fitting cyclic trends. This comparison qualitatively validates \method's superior dual adaptability in capturing both high-frequency disturbances and long-term periodic patterns. Complementing the point forecasts, \autoref{fig:quantile_prediction_vis} visualizes the robust probabilistic modeling capabilities. The results show that the ground truth trajectories reliably consistently fall within the predicted 99\% and 80\% confidence intervals. Notably, the interval widths differ dynamically adaptively—expanding during volatile fluctuations and narrowing during stable trends—demonstrating that \method effectively characterizes the inherent and dynamic aleatoric uncertainty inherent in traffic series.

\subsection{Efficiency Analysis}
\begin{figure}[h!]
    \centering
    \includegraphics[width=\columnwidth]{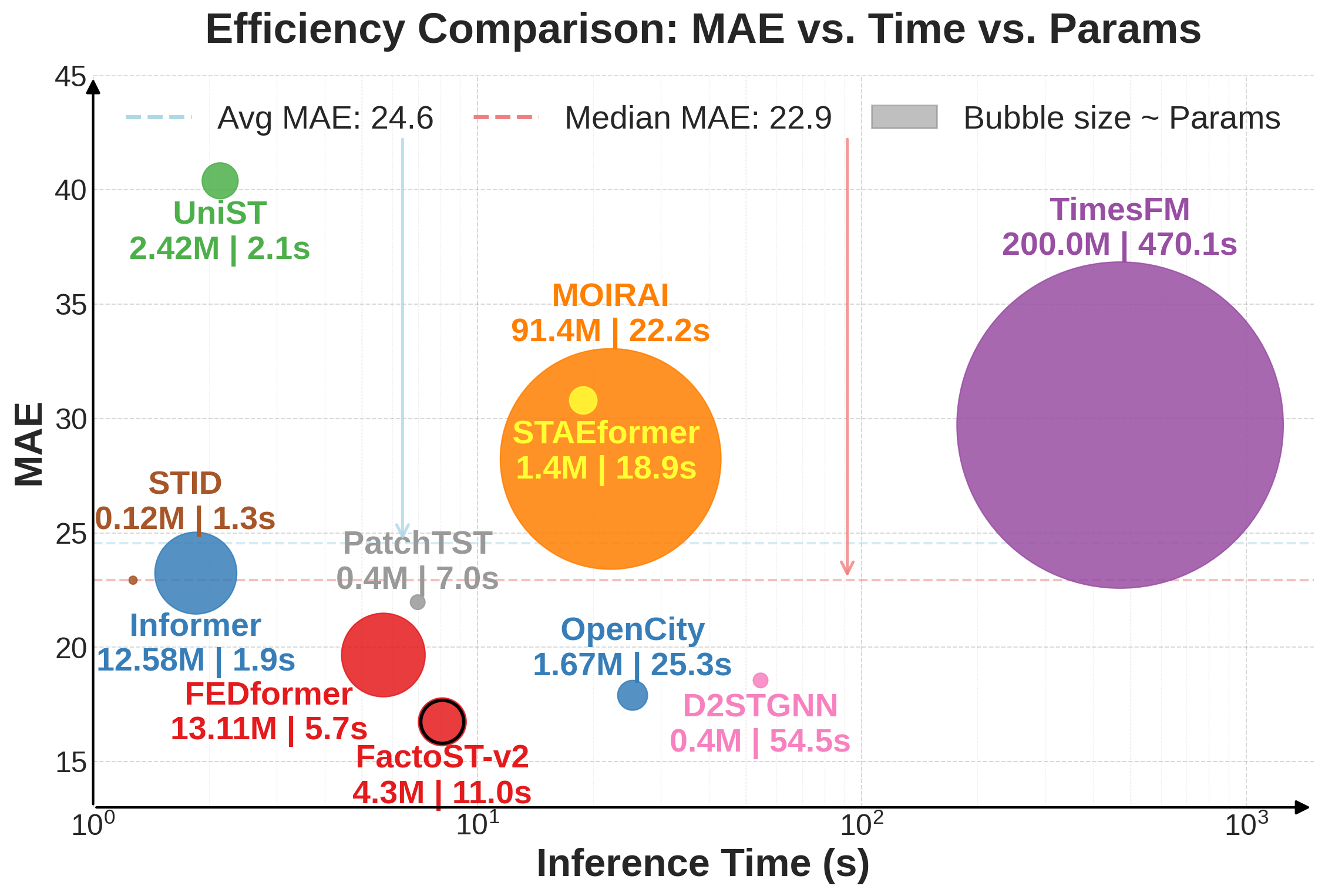}
    \caption{Accuracy-efficiency trade-off: MAE vs. inference time vs. parameters. Each point is a model; bubble size = parameter scale. \method (red) lies in Pareto-optimal region (low MAE/latency, moderate parameters)}
    \label{fig:efficiency}
\end{figure}

As shown in \autoref{fig:efficiency}, \method achieves a consistently strong MAE of 16.65 on PEMS-03 under the 10\% few-shot setting with merely 4.3M parameters and 11.0s inference time—vastly outperforming nearly all baselines in both accuracy and efficiency. In contrast, joint ST models like OpenCity (1.67M params, 25.3s) notably incur high computational overhead due to end-to-end graph learning, while large-scale TSFMs such as Moirai (91.4M params, 22.2s) ignore spatial structure entirely. Even though models like D2STGNN attain slightly better accuracy, they suffer from prohibitively severe latency (54.5s). By decoupling universal temporal pretraining from lightweight, plug-and-play spatial adaptation, \method achieves an exceptional \textit{accuracy–efficiency trade-off}, making it highly suitable for real-world deployment.

\section{Conclusion}
\label{conclusion}

This work validates the \textit{Pattern Factorization Hypothesis} for ST Foundation Models. Instead of pursuing monolithic architectures that entangle complex dynamics, we demonstrated that \textit{decoupling universal temporal laws from domain-specific spatial contexts} offers a more scalable and effective path toward general intelligence in physical systems. Our empirical findings yield three critical insights for the community:

\begin{enumerate}[leftmargin=*]
    \item \textbf{Universality lies in Time, not Space:} We observe that temporal patterns possess high transferability across disparate domains, whereas spatial dependencies are inherently task-specific. Isolating the former allows models to scale effectively on billions of observations without being hampered by topological conflicts (negative transfer).
    \item \textbf{Efficiency dictates Scalability:} The success of our lightweight adaptation mechanism challenges the prevailing trend of heavy fine-tuning. It proves that a frozen, robust temporal backbone can be efficiently steered toward complex spatial tasks via minimal parameter injection, breaking the quadratic scalability wall.
    \item \textbf{Unified Regimes:} The factorized paradigm bridges the performance gap across diverse data regimes, showing that a single backbone can simultaneously excel in zero-shot transfer, few-shot adaptation, and full supervised learning.
\end{enumerate}

Looking forward, \method paves the way for \textbf{democratizing STFMs}. By reducing the computational barrier of adaptation and enabling the \textbf{100\% reuse} of powerful pretrained temporal knowledge for ST tasks, it opens new avenues for applying foundation models to resource-constrained scientific domains. We hope this work encourages future research to further explore the synergy between general-purpose sequential modeling and domain-specific structural priors.

\section{Limitations and Future Work}
\label{sec:limitations}
While \method establishes a robust factorized paradigm for scalable ST modeling, it also points out several open challenges that pave the way for future exploration.

\begin{itemize}[leftmargin=*]
    \item \textbf{Towards Open-World Spatial Generalization.} Although our lightweight adapter efficiently handles domain shifts, the current framework largely operates under a transductive or semi-inductive setting, relying on learned embeddings for specific nodes. A critical next step is to achieve \textit{fully inductive, zero-shot spatial generalization} on dynamic, open-world topologies where node sets evolve continuously \cite{ma2025beyond,marobust,wu2025spatio}. Future work could explore \textit{Text-to-Space alignment}—leveraging Large Language Models (LLMs) to generate semantic graph structures from auxiliary metadata (e.g., POI descriptions) without training on historical data~\cite{chen2025enhancing,feng2025citybench,he2025geolocation,liu2024time}. Additionally, integrating \textit{test-time adaptation} ~\cite{guo2025online,chen2024test,chen2025learning, liu2024online, liu2024refol} to dynamically infer latent relational structures from streaming data would further enhance \method's robustness against non-stationary graph evolution~\cite{lei2025st,zhou2025coms2t}.

    \item \textbf{Unified Modeling of Exogenous Causality.} Our factorized framework inherently provides a modular and plug-and-play architecture for extensions. Currently, like most STFMs, \method primarily models endogenous historical patterns. However, real-world dynamics are often driven by rich, multi-modal exogenous factors (e.g., events) \cite{siru2025time,zheng2025fusing,zhan2025time}. Designing a unified, lightweight interface to fuse such signals into our factorized framework—potentially via \textit{LLM-powered covariate encoders} or universal cross-modal adapters~\cite{wang2024timexer}—would be a natural and valuable extension, enhancing its applicability and causal reasoning capacity.
\end{itemize}

\section{Acknowledgments}
\noindent This work was mainly supported by the National Natural Science Foundation of China (Grant No. 62402414); Huawei (Grant No. TC20241023027); the Guangdong Basic and Applied Basic Research Foundation (Grant No. 2025A1515011994); the Guangzhou Municipal Science and Technology Project (Grant No. 2023A03J0011); and the Guangzhou Industrial Information and Intelligent Key Laboratory Project (Grant No. 2024A03J0628).

\bibliographystyle{IEEEtran}
\bibliography{main}

\vspace{11pt}

\vfill

\end{document}